\documentclass[10pt]{article} %
\usepackage[accepted]{tmlr}

\usepackage{amsmath,amsfonts,bm}

\def\eqref#1{equation~\ref{#1}}

\def\1{\bm{1}}

\DeclareMathAlphabet{\mathsfit}{\encodingdefault}{\sfdefault}{m}{sl}
\SetMathAlphabet{\mathsfit}{bold}{\encodingdefault}{\sfdefault}{bx}{n}

\usepackage{bbding}
\usepackage{bbding}
\usepackage{rotating}
\usepackage{epsfig}
\usepackage{graphicx}
\usepackage{amsmath}
\usepackage{amssymb}
\usepackage{comment}
\usepackage{threeparttable}

\usepackage[utf8]{inputenc} %
\usepackage[T1]{fontenc}    %
\usepackage{url}            %
\usepackage{booktabs}       %
\usepackage{amsfonts}       %
\usepackage{nicefrac}       %
\usepackage{microtype}      %
\usepackage{xcolor}         %
\usepackage{colortbl}

\usepackage{multirow}
\usepackage{enumitem}
\usepackage{makecell}
\usepackage{pifont}

\usepackage{arydshln}
\usepackage{bbm}
\usepackage[export]{adjustbox}
\usepackage{wrapfig}

\usepackage{pifont} %
\newcommand{\cmark}{\ding{51}} %
\newcommand{\xmark}{\ding{55}} %

\definecolor{LightGray}{rgb}{0.9,0.9,0.9}
\definecolor{DarkRed}{rgb}{0.75,0,0}
\definecolor{DarkBlue}{rgb}{0,0,0.55}
\definecolor{DarkGreen}{rgb}{0.43, 0.68, 0.28}

\definecolor{xishen}{rgb}{0.858, 0.188, 0.478}

\definecolor{DarkBlue}{rgb}{0.0, 0.5, 0.8}

\usepackage{hyperref}

\usepackage{url}
\usepackage{amsmath}
\usepackage{wasysym}
\usepackage{booktabs}
\usepackage{adjustbox}
\usepackage{float}
\usepackage{placeins}
\usepackage{pifont}
\usepackage{wrapfig}

\usepackage{subcaption} 
\usepackage{amssymb}

\usepackage{tabu}

\definecolor{LightGray}{rgb}{0.9,0.9,0.9}
\title{\smash{\raisebox{-0.2\height}{\includegraphics[height=0.9cm]{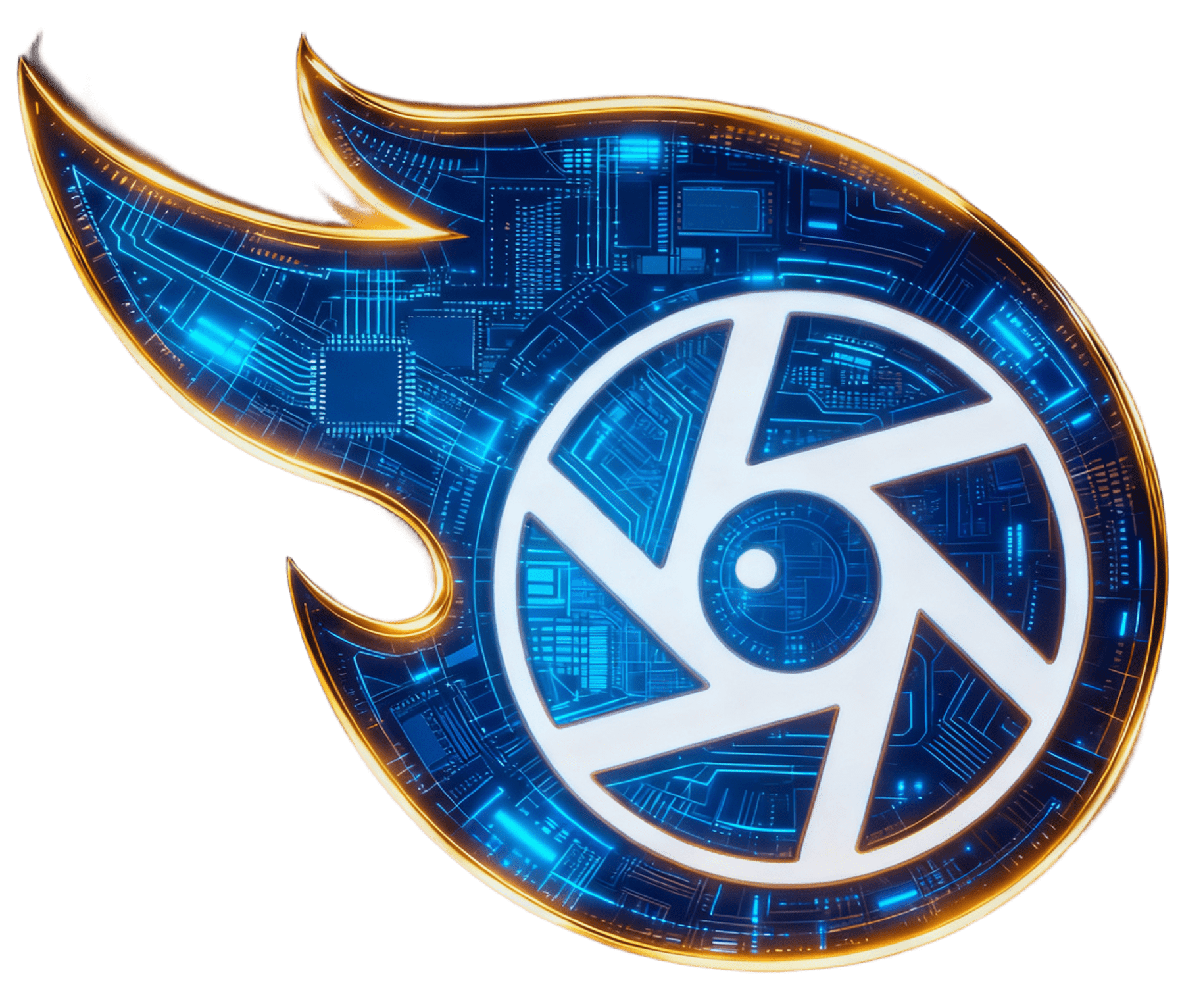}}}\hspace{0.15em}EdgeCrafter: Compact ViTs for Edge Dense Prediction via Task-Specialized Distillation}

\author{\name Longfei Liu$^{*\ddagger}$, 
Yongjie Hou$^{*}$, 
Yang Li$^{*}$, 
Qirui Wang$^{*}$, 
Youyang Sha, 
Yongjun Yu, 
Yinzhi Wang,  
Peizhe Ru, 
Xuanlong Yu$^{\dagger}$, 
Xi Shen$^{\dagger}$ \\
\\
Intellindust AI Lab \\
\\
{\small $^{*}$ Equal contribution \quad
$^{\ddagger}$ Project lead \quad
$^{\dagger}$ Corresponding authors}
}

\newcommand{\edgecrafter}{EdgeCrafter}

\newcommand{\detector}{ECDet}
\newcommand{\seg}{ECInsSeg}
\newcommand{\pose}{ECPose}

\def\month{08}  %
\def\year{2026} %
\def\openreview{\url{https://openreview.net/forum?id=yoSD5BRFkV}} %

\begin{document}

\maketitle

\begin{abstract}
    Deploying high performance dense prediction models on resource-constrained edge devices remains challenging due to strict limits on computation and memory. In practice, lightweight systems for object detection, instance segmentation, and pose estimation are still dominated by CNN-based architectures such as YOLO, while compact Vision Transformers (ViTs) often struggle to achieve similarly strong accuracy--efficiency trade-offs, even with large scale pretraining. We argue that this gap is largely due to insufficient task-specific representation learning in small-scale ViTs, rather than an inherent mismatch between ViTs and edge dense prediction. To address this issue, we introduce \edgecrafter, a unified compact ViT framework for edge dense prediction centered on \detector, a detection model built from a distilled compact backbone and an edge-friendly encoder--decoder design. We first adapt a large DINOv3-pretrained ViT to object detection and use it as a task-specialized teacher to distill rich representations into compact student backbones on ImageNet-1K and COCO images. We further improve efficiency by replacing standard patch embedding with a lightweight convolutional stem and constructing multi-scale features with simple interpolation and linear projection instead of costly feature pyramids. The resulting detection-distilled representation transfers directly to instance segmentation and human pose estimation through lightweight task-specific prediction modules. Using no task annotations beyond COCO, \detector-S reaches 51.7 box AP with fewer than 10M parameters, while \seg-X and \pose-X reach 48.4 mask AP and 74.8 keypoint AP. As a complementary but more compute-intensive setting, Objects365 detection pretraining consistently improves all scales, with the X variants reaching 59.9 box AP, 49.8 mask AP, and 75.9 keypoint AP. These results show that compact ViTs, when paired with task-specialized distillation and edge-aware design, can be a practical and competitive option for edge dense prediction. Code is available at:~\url{https://intellindust-ai-lab.github.io/projects/EdgeCrafter/}
    
\end{abstract}

\section{Introduction}

\begin{figure*}[t]
\centering

\begin{subfigure}{0.32\linewidth}
  \centering
  \includegraphics[width=\linewidth]{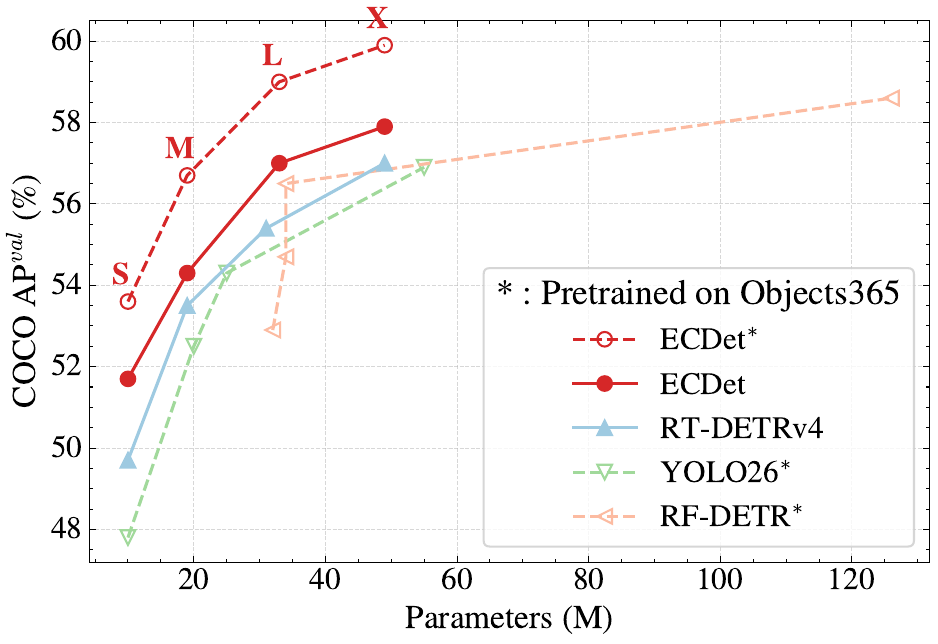}
\end{subfigure}%
\hfill
\begin{subfigure}{0.32\linewidth}
  \centering
  \includegraphics[width=\linewidth]{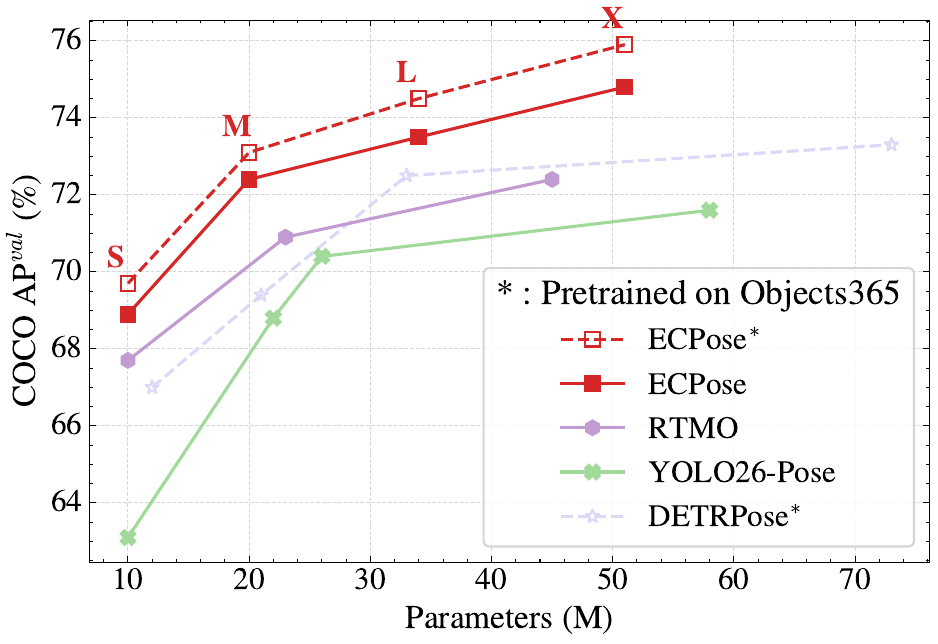}
\end{subfigure}%
\hfill
\begin{subfigure}{0.32\linewidth}
  \centering
  \includegraphics[width=\linewidth]{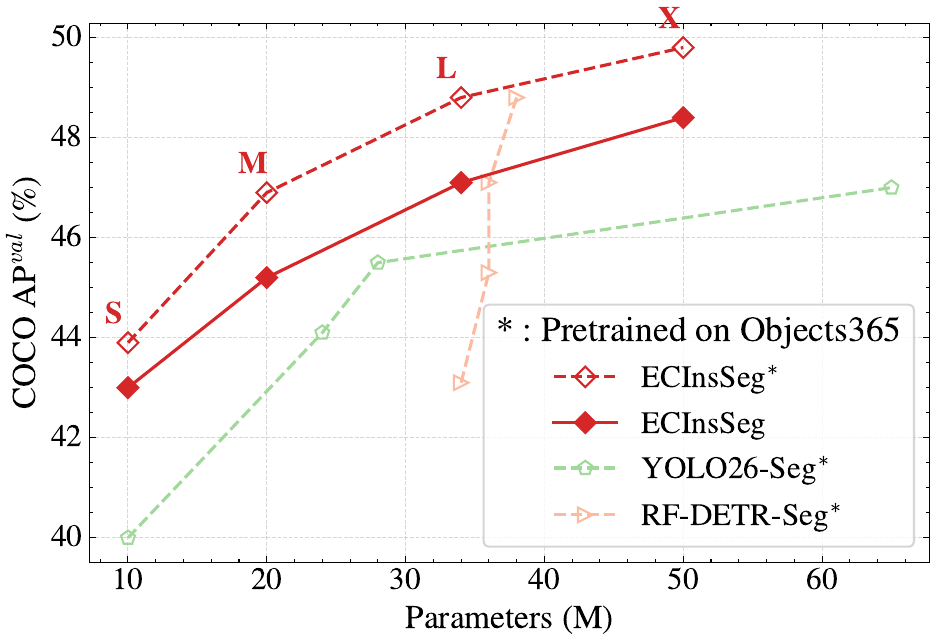}
\end{subfigure}

\begin{subfigure}{0.32\linewidth}
  \centering
  \includegraphics[width=\linewidth]{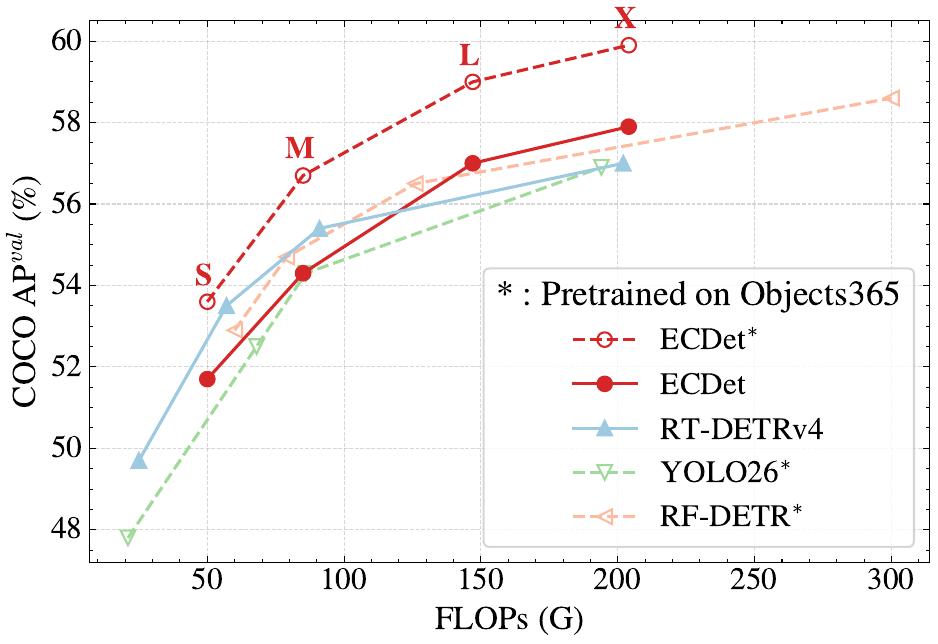}
  \subcaption*{\textbf{Object Detection}}
\end{subfigure}%
\hfill
\begin{subfigure}{0.32\linewidth}
  \centering
  \includegraphics[width=\linewidth]{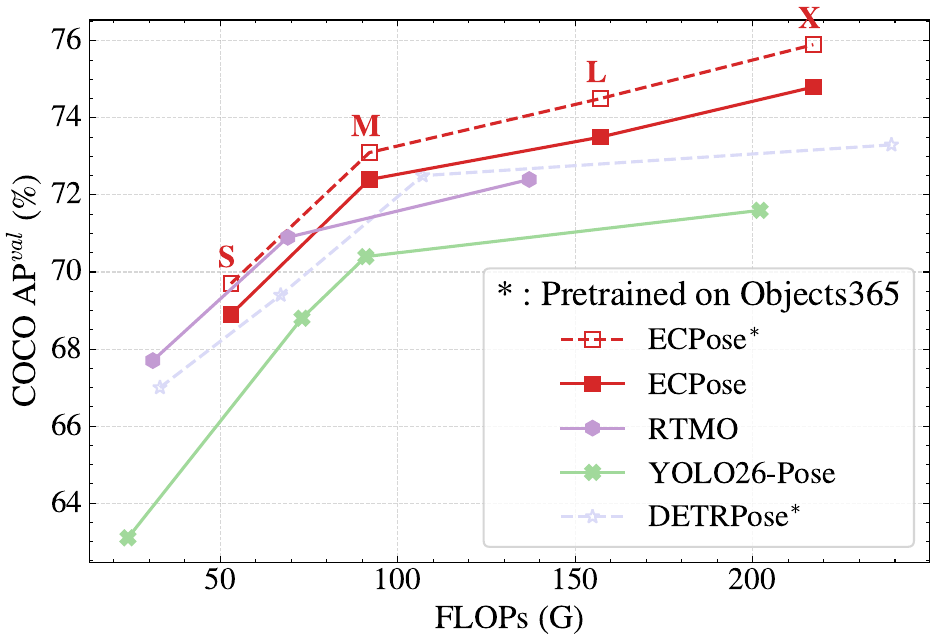}
  \subcaption*{\textbf{Human Pose Estimation}}
\end{subfigure}%
\hfill
\begin{subfigure}{0.32\linewidth}
  \centering
  \includegraphics[width=\linewidth]{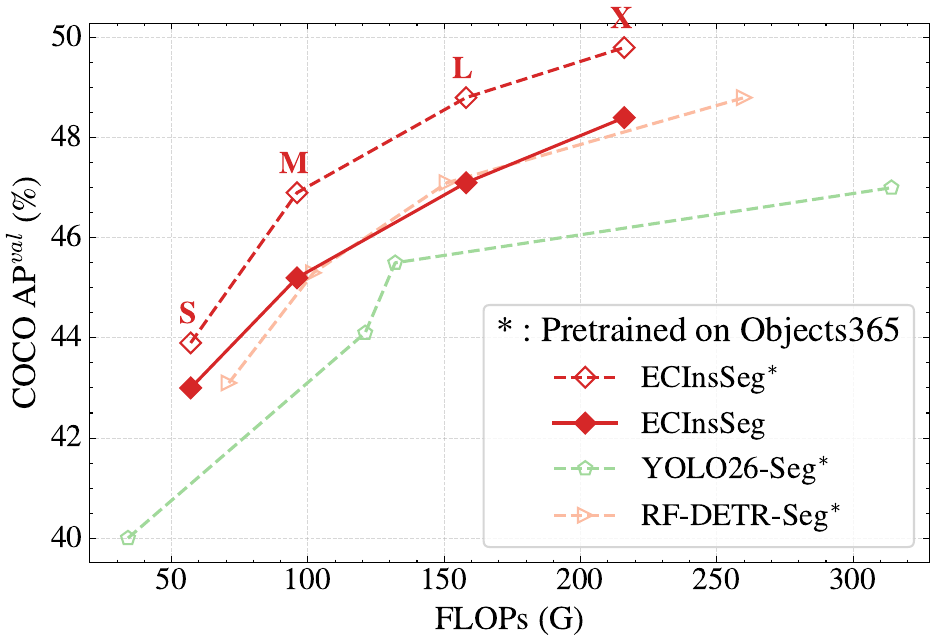}
  \subcaption*{\textbf{Instance Segmentation}}
\end{subfigure}

\begin{subfigure}{\linewidth}
\centering
\phantomsubcaption
\subcaption*{\textbf{(a) Compared with SOTA across dense prediction tasks on COCO}}
\label{fig:teaser:sota}
\end{subfigure}
\vspace{6pt}

\begin{subfigure}{0.42\linewidth}
  \centering
  \includegraphics[width=\linewidth]{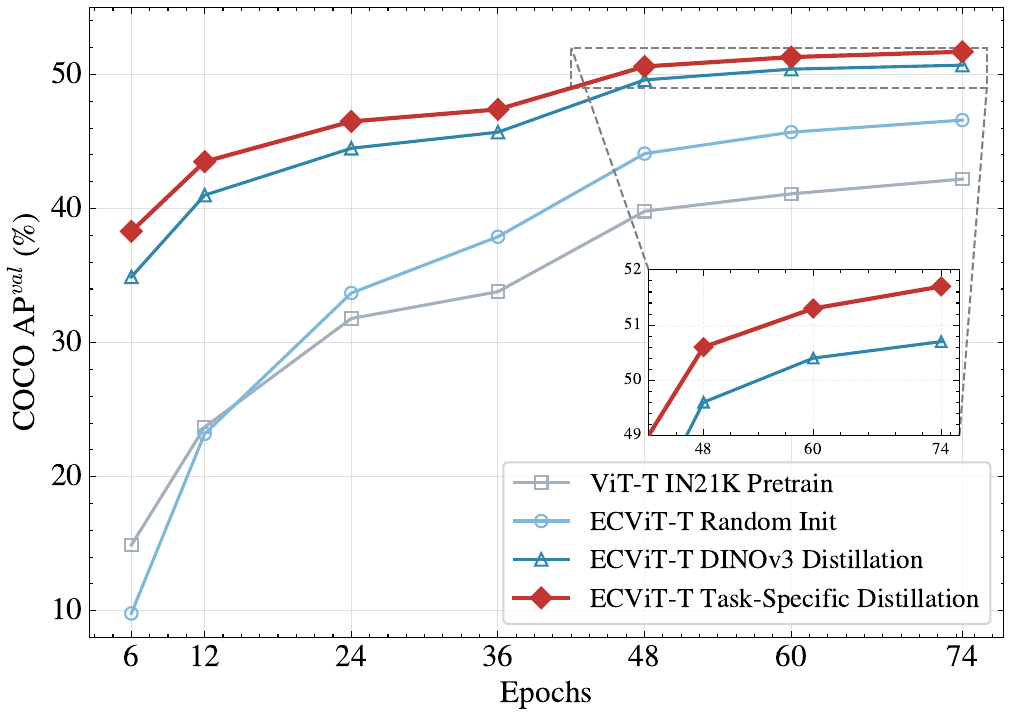}
\end{subfigure}%
\hfill
\begin{subfigure}{0.50\linewidth}
  \centering
  \includegraphics[width=\linewidth]{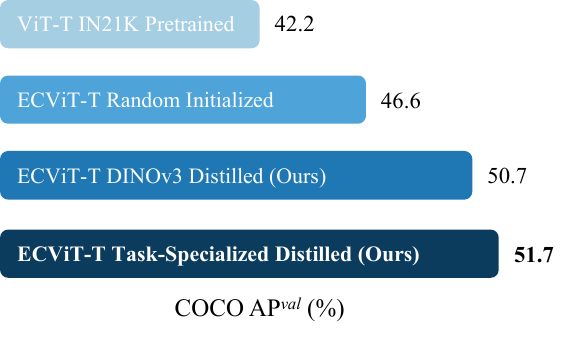}
\end{subfigure}

\begin{subfigure}{\linewidth}
\centering
\phantomsubcaption
\subcaption*{\textbf{(b) Effect of pretraining and task-specialized distillation}}
\label{fig:teaser:pretrain}
\end{subfigure}

\caption{\textbf{Comprehensive evaluation of EdgeCrafter.}
(a) Accuracy--efficiency comparison across object detection, human pose estimation, and instance segmentation on COCO~\citep{lin2014microsoft}. The top and bottom rows plot model parameters and FLOPs, respectively, against AP. Methods marked with $^{*}$ use Objects365 pretraining~\citep{shao2019objects365}. (b) Training dynamics and downstream AP for ECDet-T under different backbone initialization strategies. ViT-T follows the ImageNet-21K~\citep{ridnik2021imagenet} training strategy of~\citet{steiner2022train}. Supervised ImageNet-21K pretraining is weaker than training from scratch for this compact model, while task-specialized distillation yields substantially stronger downstream performance.}
\label{fig:teaser}
\end{figure*}

Object detection~\citep{girshick2015fast}, instance segmentation~\citep{He_2017_ICCV}, and human pose estimation~\citep{cao2019openpose} are core dense prediction tasks in computer vision, with applications ranging from autonomous driving~\citep{arnold2019survey} and robotics~\citep{labbe2020cosypose} to augmented reality~\citep{chan2019everybody} and video analytics~\citep{caetano2019skelemotion}. Bringing these models to edge devices is attractive for reasons such as responsiveness and privacy, but remains challenging due to the diverse resource and deployment constraints of edge platforms. In practice, edge deployment requires a careful balance between predictive performance and deployment efficiency, involving not only computational cost and model size, but also memory requirements and real-time inference latency. The relative importance of these factors can further vary across hardware and software platforms. These constraints are especially pronounced in battery-powered or embedded settings, where resource efficiency directly affects the practicality of deploying dense prediction models.

In real-time edge deployment scenarios, dense prediction tasks are still overwhelmingly dominated by models built upon convolutional neural network (CNN) backbones. The YOLO family~\citep{redmon2016you,redmon2017yolo9000,yolov8,yolo11,li2022yolov6,wang2024yolov9,wang2024yolov10} and DETR-based models~\citep{lv2024rt, peng2025dfine} are representative examples, offering strong accuracy–efficiency trade-offs. While pioneering attempts have explored adopting Vision Transformers (ViTs) as backbones for dense prediction~\citep{chen2024lw,li2022exploring}, these approaches typically rely on scaling to large model sizes (e.g., ViT-Base or Large) or require additional massive pretraining datasets such as Objects365~\citep{shao2019objects365} to achieve competitive performance. Recently, large-scale self-supervised pretraining paradigms~\citep{oquab2023dinov2,simeoni2025dinov3} have unlocked the extraordinary representation capabilities of ViTs. %
Recent works such as RF-DETR~\citep{robinson2025rf} directly employ DINOv2~\citep{oquab2023dinov2} as a backbone. However, even the smallest variant contains approximately 30M parameters, imposing prohibitive computational and memory constraints for strict edge deployment. %

Unfortunately, naively shrinking a ViT to meet stringent edge constraints results in a precipitous drop in performance. %
Our empirical observations show that even initializing compact ViTs with supervised IN-21K~\citep{ridnik2021imagenet,steiner2022train} pretraining~\footnote{\url{https://huggingface.co/timm/vit_tiny_patch16_224.augreg_in21k}} yields unsatisfactory performance for dense prediction, sometimes even performing worse than training from scratch (Figure~\ref{fig:teaser:pretrain}). This observation, which is consistent with findings reported by~\citet{ghiasi2020simple, zoph20selftraining}, suggests that generic supervised pretraining alone is often insufficient for compact ViTs in edge dense prediction. %

In this work, we introduce \textit{\edgecrafter}, a unified compact ViT framework for edge dense prediction centered on \detector. Our approach is built upon two key ideas. First, we employ task-specialized knowledge distillation: a large DINOv3-pretrained ViT~\citep{simeoni2025dinov3} is first adapted for object detection and then used as a teacher to supervise compact student models. Second, we design an edge-friendly student architecture that replaces the standard patch embedding with a lightweight convolutional stem~\citep{xiao2021early} and constructs multi-scale features using simple interpolation and linear projections. The resulting distilled backbone and encoder are used directly in \detector\ and can be transferred to \seg\ and \pose\ through lightweight task-specific heads, resulting in a unified framework for object detection, instance segmentation, and human pose estimation.

Figure~\ref{fig:teaser} summarizes the two main messages of this work. First, Figure~\ref{fig:teaser:sota} shows that \edgecrafter~achieves strong accuracy-efficiency trade-offs across object detection, instance segmentation, and human pose estimation, measured by parameters and FLOPs, while reusing a common detection-distilled representation. Several competitive baselines rely on Objects365 detection pretraining~\citep{shao2019objects365}, yet \edgecrafter~remains competitive in our primary non-O365 setting, which uses no task annotations beyond COCO while retaining the DINOv3-based distillation described in Section~\ref{sec:method_distillation}. We report O365 as a complementary, more compute-intensive setting that consistently improves all scales. Second, Figure~\ref{fig:teaser:pretrain} shows that the representation learning strategy matters substantially: standard generic pretraining is not enough for compact ViTs, whereas task-specialized distillation leads to much stronger downstream performance. Together, these results support the view that compact ViTs can be competitive for edge dense prediction when their representations and architecture are designed for the target setting.

In summary, this work makes the following contributions:

\begin{itemize}
    \item We identify insufficient task-specific representation learning as a key bottleneck for compact ViTs in edge dense prediction, and show that generic supervised pretraining alone is often inadequate for small models.
    \item We propose an edge-oriented compact ViT design that combines task-specialized distillation, a lightweight convolutional stem, and simple multi-scale feature construction, making ViT backbones more suitable for dense prediction under tight parameter and FLOP budgets.
    \item We introduce \edgecrafter, a unified framework centered on \detector, and show that its detection-distilled representation transfers effectively to instance segmentation and human pose estimation while maintaining strong cross-task accuracy-efficiency trade-offs.
\end{itemize}

Code is available at:~\url{https://intellindust-ai-lab.github.io/projects/EdgeCrafter/}.

\section{Related Work}

\paragraph{Knowledge distillation from vision foundation models.}
Knowledge distillation (KD)~\citep{hinton2015distilling,kim2018paraphrasing,ba2014deep,mirzadeh2020improved,beyer2022knowledge} is a standard tool for transferring knowledge from large models to smaller ones, either through logits or intermediate features~\citep{Romero2015FitNetsHF,huang2017like,ahn2019variational,heo2019overhaul,zagoruyko2017paying,wei2022featuredistillation}. With the emergence of vision foundation models, recent work has explored distillation from CLIP~\citep{radford2021learning}, SAM~\citep{kirillov2023segment}, DINOv2~\citep{oquab2023dinov2}, and DINOv3~\citep{simeoni2025dinov3} into smaller backbones or unified representations~\citep{wei2022featuredistillation,wang2024sam,ranzinger2024radio}. These studies show that powerful pretrained teachers can improve downstream transfer, but they are mostly designed for general-purpose representation learning rather than compact edge-oriented dense prediction. In contrast, our goal is to convert a large pretrained ViT into a task-specialized teacher for dense prediction, and then distill that knowledge into compact students that remain effective under tight parameter and FLOP budgets.

\paragraph{Efficient object detection.}
Efficient object detectors are still dominated in practice by CNN-based families such as YOLO~\citep{redmon2016you,redmon2017yolo9000,redmon2018yolov3,li2022yolov6,wang2024yolov9,wang2024yolov10,yolov8,yolo11,tian2025yolov12}, which provide strong accuracy-efficiency trade-offs and broad task coverage. DETR~\citep{carion2020detr} introduced end-to-end set prediction for detection, and subsequent work improved convergence and efficiency~\citep{zhu2021deformable,liu2022dab,li2022dn,zhang2023dino,zhao2024detrs,lv2024rt,peng2025dfine,wang2025rt}. More recent detectors have combined this line of work with ViT backbones and large-scale pretraining. LW-DETR~\citep{chen2024lw} uses a pretrained ViT backbone, RF-DETR~\citep{robinson2025rf} further incorporates DINOv2~\citep{oquab2023dinov2} and architecture search, DEIMv2~\citep{huang2025deimv2} combines DINOv3~\citep{simeoni2025dinov3} with lightweight convolutions, and RT-DETRv4~\citep{liao2025rtdetrv4} aligns detector training with DINOv3 representations. These works demonstrate the value of pretrained transformer features, but strong performance often depends on comparatively large backbones or additional large-scale pretraining. Our work instead focuses on transferring task-specialized ViT representations into compact students for edge dense prediction.

\paragraph{Efficient instance segmentation.}
Instance segmentation methods have evolved from two-stage frameworks such as Mask R-CNN~\citep{He_2017_ICCV} to more efficient one-stage or end-to-end designs, including YOLACT~\citep{bolya2019yolact}, BlendMask~\citep{chen2020blendmask}, SOLO~\citep{wang2020solo}, and FastInst~\citep{wu2023fastinst}. In parallel, query-based transformer methods such as MaskFormer~\citep{cheng2021maskformer}, Mask2Former~\citep{cheng2022mask2former}, and MaskDINO~\citep{li2023maskdino} have shown that end-to-end segmentation can benefit from strong dense representations. RF-DETR-Seg~\citep{robinson2025rf} further extends the pretrained RF-DETR framework to efficient instance segmentation. Relative to this line of work, our goal is not to introduce a new segmentation-specific backbone, but to show that a compact distilled ViT backbone can support instance segmentation effectively within the same unified framework used for detection and pose estimation.

\paragraph{Efficient human pose estimation.}
Human pose estimation methods include top-down and bottom-up pipelines, as well as more recent end-to-end set prediction approaches. Practical efficient systems often build on CNN-based or YOLO-style detectors~\citep{rtmo,yolo11,yolopose}, while DETR-based pose estimators such as PETR~\citep{petr}, QueryPose~\citep{querypose}, ED-Pose~\citep{edpose}, GroupPose~\citep{grouppose}, and DETRPose~\citep{janampa2025detrpose} adapt set-based prediction to multi-person keypoint estimation. These methods show that transformer decoders can be effective for pose estimation, especially when paired with strong detection backbones. Our work is complementary: we focus on the backbone side of the problem and show that a compact distilled ViT can serve as a shared representation for pose estimation, while also supporting object detection and instance segmentation in the same framework.

\begin{figure*}[t]
\centering
  \includegraphics[width=1.0\linewidth]{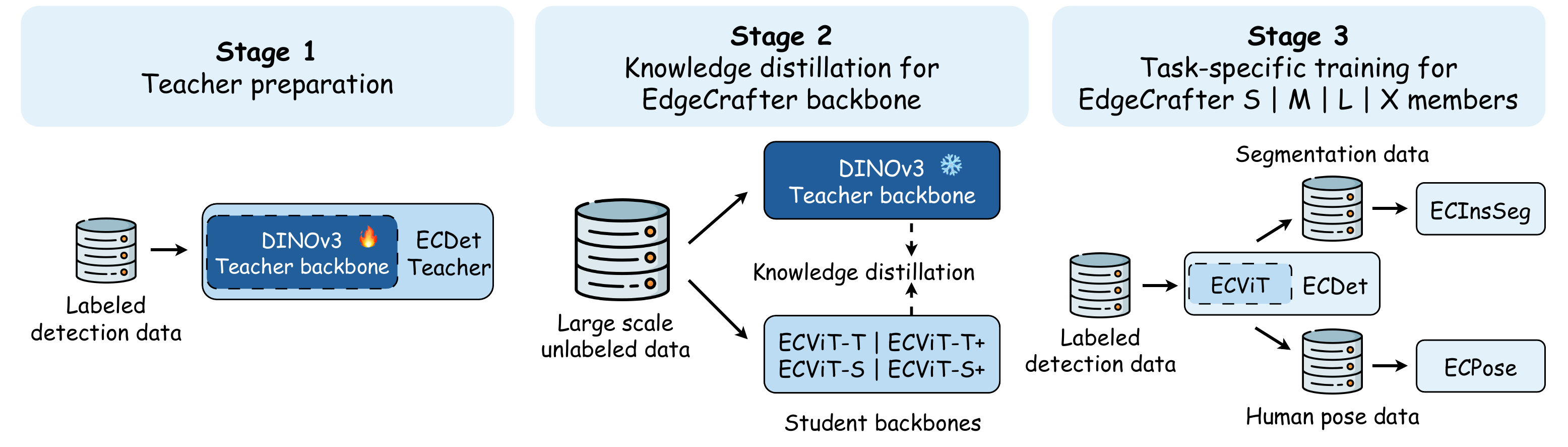}
  \caption{\textbf{Overview of the \edgecrafter~pipeline.} \textbf{Stage 1:} A pretrained DINOv3 backbone~\citep{simeoni2025dinov3} is adapted to object detection to create a task-specialized teacher within the ECDet formulation. \textbf{Stage 2:} The resulting teacher distills its detection-oriented representation into compact ECViT student backbones through feature alignment on a large image collection. \textbf{Stage 3:} The distilled students are used to instantiate the ECDet model family at different scales (S/M/L/X), and the same distilled backbone and encoder are further reused for instance segmentation and human pose estimation with lightweight task-specific heads. The key idea is that detection serves as the representation-learning stage, while the learned backbone transfers directly to other dense prediction tasks.}
  \label{fig:pipe} 
\end{figure*}

\section{Method}
Our approach is illustrated in Figure~\ref{fig:pipe}. We first adapt a pretrained DINOv3 model~\citep{simeoni2025dinov3} to object detection and use it as a task-specialized teacher. We then distill its representations into a compact ViT backbone, denoted ECViT, and build \detector~on top of the distilled backbone. Finally, we reuse the same distilled backbone and encoder in the instance segmentation model \seg\ and the human pose estimation model \pose\ by attaching lightweight task-specific heads. This yields a unified framework in which detection is the primary training setting, while the learned representation transfers to other dense prediction tasks.

\subsection{Overview}

\detector~is the main instantiation of \edgecrafter. It combines a compact ViT backbone with an RT-DETR-style encoder-decoder detector~\citep{lv2024rt}. The backbone is adapted to dense prediction in two ways: we replace the standard large-stride patch embedding with a convolutional stem that preserves local structure, and we construct a lightweight multi-scale feature pyramid from the final transformer blocks instead of using a heavy pyramid module. To make this compact backbone effective, we distill it from a detection-specialized teacher using feature alignment. The resulting representation is then reused by \seg\ and \pose, which share the same backbone and encoder but use task-specific prediction modules.

\subsection{ECDet Architecture}
\label{sec:ecdet}

\begin{figure*}[t]
\centering
    \includegraphics[width=\linewidth]{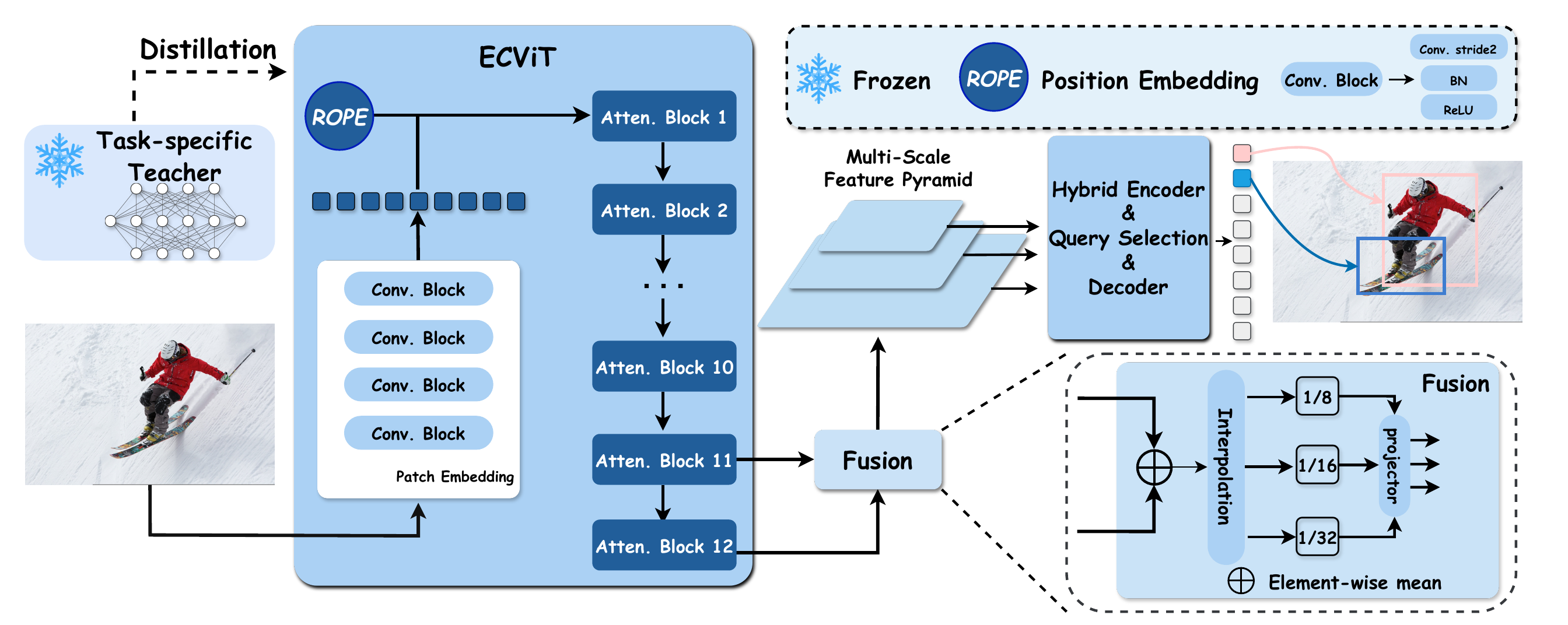}
\caption{\textbf{Architecture of ECDet.} A frozen detection-specialized teacher distills the compact ECViT backbone, whose convolutional stem replaces standard large-stride patch embedding. The final two transformer features are fused, converted into stride-$8$, stride-$16$, and stride-$32$ feature maps, refined by the hybrid encoder, and decoded into object predictions.}
\label{fig:ecdet}
\end{figure*}

\textbf{ECDet} consists of a compact ViT backbone, denoted ECViT, followed by an encoder and a decoder for set-based object prediction. Figure~\ref{fig:ecdet} summarizes the architecture.

\paragraph{Backbone design.}
Compact ViTs for detection must preserve local structure while remaining efficient. A standard ViT patch embedding applies a single large-stride projection at the input, which is effective for image classification but can discard fine spatial details that matter for dense localization. Following prior work that replaces patch embedding with a convolutional stem~\citep{xiao2021early}, we use a stack of four $3 \times 3$ convolutions with stride $2$ in the ECDet backbone. Our motivation is detection-specific: instead of aggregating local information in one step, the stem enlarges the receptive field progressively before the transformer blocks. This is also consistent with effective receptive field analyses~\citep{luo2016understanding}, which show that stacked convolutions retain a center-concentrated effective receptive field even when their theoretical receptive field grows. We therefore use a moderate receptive field in the stem and analyze this choice in the detection ablations.

\paragraph{Multi-scale feature generation.}
ViTs do not naturally produce hierarchical feature pyramids, so we construct one explicitly from the final transformer blocks. Let $\mathbf{X}_{L-1}, \mathbf{X}_L \in \mathbb{R}^{(\frac{H}{16}\frac{W}{16}) \times C}$ denote the output tokens from the last two transformer blocks. We first average and reshape them into a spatial feature map at stride $16$:
\[
\mathcal{F}^{(16)} = \frac{1}{2}\left(\mathbf{X}_{L-1} + \mathbf{X}_L\right),
\]
and then form a three-level pyramid $\{\mathcal{F}^{(8)}, \mathcal{F}^{(16)}, \mathcal{F}^{(32)}\}$ with bilinear interpolation $\mathcal{B}_s(\cdot)$ and a $1 \times 1$ convolutional projection $\Theta_s(\cdot)$:
\[
\mathcal{F}^{(s)} = \Theta_s \big( \mathcal{B}_s(\mathcal{F}^{(16)}) \big), \quad s \in \{8, 16, 32\}.
\]
This design is deliberately simple and introduces very few additional parameters beyond the backbone itself, which is desirable for edge dense prediction. At the same time, it supplies the detector with the multi-scale features needed for dense localization without resorting to heavier feature pyramid modules.

\paragraph{Encoder.}
Our encoder follows RT-DETR~\citep{lv2024rt}. The coarsest feature map is first refined by attention-based intra-scale feature interaction (AIFI)~\citep{lv2024rt}, which applies self-attention on the lowest-resolution feature map to enlarge its receptive field and strengthen long-range contextual interactions at modest cost:
\[
\hat{\mathcal{F}}^{(32)} = \mathrm{AIFI}\big(\mathcal{F}^{(32)}\big).
\]
The refined coarse feature is then fused with the higher-resolution features through CNN-based cross-scale feature fusion (CCFF)~\citep{lv2024rt}, which propagates the high-level context from $\hat{\mathcal{F}}^{(32)}$ back to the finer scales and combines them with local spatial detail:
\[
\mathcal{F}_{\mathrm{enc}} = \mathrm{CCFF}\big(\hat{\mathcal{F}}^{(32)}, \mathcal{F}^{(16)}, \mathcal{F}^{(8)}\big).
\]
The encoded representation $\mathcal{F}_{\mathrm{enc}}$ provides the shared dense representation used by ECDet and by the lightweight extensions to segmentation and pose.

\FloatBarrier
\paragraph{Decoder.}
The decoder follows the DETR set-prediction formulation~\citep{carion2020detr}, with stacked self-attention, deformable cross-attention~\citep{zhu2021deformable}, and feed-forward layers operating on learned object queries $\mathbf{Q} \in \mathbb{R}^{N \times C}$. We use four decoder layers for all ECDet variants and fix the number of queries to $N=300$, which keeps the detector compact while preserving strong set-based prediction performance.

\paragraph{Training objective.}

\detector\ is trained with a standard DETR-style bipartite matching objective. Specifically, the total detection loss is defined as:
\[
\mathcal{L}_{\mathrm{det}} = \lambda_{\mathrm{cls}}\mathcal{L}_{\mathrm{cls}} +
\underbrace{\lambda_{\mathrm{\ell_1}}\mathcal{L}_{\ell_1} + \lambda_{\mathrm{giou}}\mathcal{L}_{\mathrm{giou}} +
\lambda_{\mathrm{ddf}}\mathcal{L}_{\mathrm{ddf}} +
\lambda_{\mathrm{fgl}}\mathcal{L}_{\mathrm{fgl}}}_\mathrm{box\ regression\ losses},
\]
where $\mathcal{L}_{\mathrm{cls}}$ denotes the classification loss~\citep{huang2025deim}. The terms $\mathcal{L}_{\ell_1}$ and $\mathcal{L}_{\mathrm{giou}}$ represent the standard $\ell_1$ distance and Generalized IoU losses~\citep{rezatofighi2019generalized} for bounding box coordinate regression. To further enhance localization precision, we incorporate the Decoupled Distillation Focal (DDF) loss $\mathcal{L}_{\mathrm{ddf}}$ and Fine-Grained Localization (FGL) loss  $\mathcal{L}_{\mathrm{fgl}}$ following the D-FINE paradigm~\citep{peng2025dfine}, which refine the box boundary distributions. The coefficients $\lambda_{\mathrm{cls}}$, $\lambda_{\ell_1}$, $\lambda_{\mathrm{giou}}$, $\lambda_{\mathrm{ddf}}$, and $\lambda_{\mathrm{fgl}}$ are hyperparameters used to balance the contribution of each task.

\paragraph{Model scaling and architecture details.} 
We instantiate \detector~at four scales, denoted \textbf{S}, \textbf{M}, \textbf{L}, and \textbf{X}, all evaluated at an input resolution of $640 \times 640$. Table~\ref{tab:modelscale} summarizes the configurations. Across this family, we keep the overall detector topology fixed and scale the model primarily through backbone and hidden dimensions. Specifically, the ECViT backbone uses four variants, T, T+, S, and S+, whose embedding dimensions increase from 192 to 384, whose number of attention heads increase from 3 to 6, and whose FFN expansion ratios increase from 4 to 6. The detector matches the encoder and decoder widths to the backbone capacity. The encoder hidden dimension grows from 192 to 256, while the encoder FFN dimension increases from 512 to 2048. The decoder always uses four layers and 300 learned object queries, but its FFN dimension is scaled from 512 to 2048 across the model family. We pair each student scale with a compatible teacher during distillation: the smaller S variant uses ECTeacher-S, while the larger M, L, and X variants use ECTeacher-B.

\begin{table*}[!t]
\caption{\textbf{ECDet model configurations.} The four detector variants share the same overall architecture and input resolution ($640 \times 640$), while scaling model capacity through the ECViT backbone width, encoder hidden dimensions, and decoder FFN dimensions. `Variant' denotes the ECViT backbone type used in each model. `Teacher' indicates the teacher backbone used during distillation. All models use a four-layer decoder and 300 object queries.}
\label{tab:modelscale}
\centering
\begin{adjustbox}{width=\textwidth}
\begin{tabular}{cc lcccc cc ccc}
\toprule
\multirow{2}{*}{Model} & \multirow{2}{*}{Resolution}
& \multicolumn{5}{c}{ECViT Backbone}
& \multicolumn{2}{c}{Encoder}
& \multicolumn{3}{c}{Decoder} \\
\cmidrule(lr){3-7}
\cmidrule(lr){8-9}
\cmidrule(lr){10-12}
 &  & Variant & Embed Dim & Attn Heads & FFN Ratio & Teacher
    & Hidden Dim & FFN Dim
    & Layers & FFN Dim & Queries \\
\midrule
S & 640 & ~~~~T  & 192 & 3 & 4 & ECTeacher-S & 192 & 512  & 4 & 512  & 300 \\
M & 640 & ~~~~T+ & 256 & 4 & 4 & ECTeacher-B & 256 & 512  & 4 & 1024 & 300 \\
L & 640 & ~~~~S  & 384 & 6 & 4 & ECTeacher-B & 256 & 1024 & 4 & 1024 & 300 \\
X & 640 & ~~~~S+ & 384 & 6 & 6 & ECTeacher-B & 256 & 2048 & 4 & 2048 & 300 \\
\bottomrule
\end{tabular}
\end{adjustbox}
\end{table*}

\subsection{Task-Specialized Distillation}
\label{sec:method_distillation}

\paragraph{Detection-specialized teacher.}
We build the teacher by adapting a pretrained DINOv3 model~\citep{simeoni2025dinov3} to object detection with the same detector formulation used by \detector. This converts a general-purpose foundation model into a detection-specialized teacher whose representation is directly aligned with the downstream student task. In the final model family, we use two teacher scales: ECTeacher-S and ECTeacher-B, obtained by adapting DINOv3-S and DINOv3-B, respectively. The details of ECTeacher are provided in Section~\ref{sec:disdetails}. These teacher scales provide a practical balance between teacher strength and student-teacher compatibility, and the influence of teacher scale is analyzed in the ablations.

\paragraph{Feature-alignment distillation.}
The student backbone is trained to match the teacher representation through one-to-many feature alignment. Let $\mathbf{X}^{S}_{L}$ denote the token features from the final student transformer block, and let $\mathbf{X}^{T}_{L-1}$ and $\mathbf{X}^{T}_{L}$ denote the token features from the last two teacher blocks. We attach a linear adapter $\phi(\cdot)$, implemented as a single learned linear layer applied token-wise, to map the student features to the teacher feature dimension. We then use the same adapted student representation to match both late teacher features:
\[
\mathcal{L}_{\mathrm{distill}} = \sum_{l \in \{L-1, L\}} \left\| \phi\!\left(\mathbf{X}^{S}_{L}\right) - \mathbf{X}^{T}_{l} \right\|_{2}^{2}.
\]
This objective is deliberately simple. It uses a single late student feature to align with multiple late teacher features, which encourages the compact student to absorb the teacher's high-level representation without introducing a heavy projection head. Using a minimal adapter keeps most of the representational burden on the student backbone itself, rather than allowing a high-capacity projection head to absorb the mismatch. We study the effect of the aligned feature depth in the ablations and use the last two teacher blocks in the final recipe.

\paragraph{Distillation setup.}
Distillation is performed before downstream task adaptation. During this stage, the compact student is optimized only to match the detection-specialized teacher using $\mathcal{L}_{\mathrm{distill}}$. The detailed training setups, including optimizer, data augmentation and training batch size, are provided in Section~\ref{sec:disdetails}.
The smaller S detector is distilled from ECTeacher-S, while the larger M, L, and X detectors are distilled from ECTeacher-B, reflecting our finding that the teacher should be matched to the student capacity rather than made arbitrarily large. We also distill with both ImageNet-1K~\citep{deng2009imagenet} and COCO~\citep{lin2014microsoft} training set images, and further adapt the teacher on COCO detection before distillation, which makes the teacher more task-specific. The effects of aligned feature depth, teacher scale, optimizer choice, register tokens, distillation dataset, and teacher adaptation are all analyzed in the ablations. After distillation, the resulting backbone is reused for object detection. The backbone trained for object detection is then further transferred to instance segmentation and pose estimation. Downstream task optimization uses only the corresponding COCO task annotations~\citep{lin2014microsoft}; this does not exclude the DINOv3 teacher supervision and feature distillation on ImageNet-1K and COCO images described above.

\subsection{Extensions to Human Pose Estimation and Instance Segmentation}
\label{sec:pose_seg}
\pose~and \seg~reuse the distilled \detector~backbone and encoder, while adopting lightweight task-specific prediction modules. In all three tasks, the task-specific training stage uses only COCO annotations~\citep{lin2014microsoft}.

\subsubsection{ECPose}

\paragraph{Pose decoder.}
ECPose formulates human pose estimation as an end-to-end set prediction problem built on top of the distilled ECDet representation. The distilled backbone and encoder are kept unchanged, and only the detection head is replaced with a lightweight pose head. Following DETRPose~\citep{janampa2025detrpose}, the decoder maintains a fixed set of learned person queries, where each query is represented by one instance token and $K$ keypoint tokens corresponding to the $K$ human joints. This structured query design makes the pose layout explicit while preserving the same set-prediction interface as ECDet.

Each decoder layer updates these query tokens through self-attention and deformable cross-attention. Self-attention first exchanges information within each person query, allowing the instance token and keypoint tokens to reason jointly about a single pose. It also allows tokens of the same keypoint type across different queries to interact, which helps suppress duplicate predictions in crowded scenes. Deformable cross-attention then samples the encoded multi-scale features around the current reference points to refine the person and keypoint locations. The final decoder outputs are used to predict one person instance together with its keypoints.

\newcommand{\ecdetresultstable}{%
\clearpage
\thispagestyle{fancy}
\noindent\begin{minipage}{\textwidth}
\captionsetup{hypcap=false}
	\captionof{table}{\textbf{Performance comparison of real-time object detectors on COCO~\citep{lin2014microsoft} \texttt{val2017}.} 
* denotes the actual training epochs, including additional epochs without augmentation. 
Latency is measured on an NVIDIA T4 GPU with batch size 1 under FP16 using TensorRT (v10.6), following RT-DETR~\citep{lv2024rt} and D-FINE~\citep{peng2025dfine}. 
YOLO latency includes NMS. 
\textcolor{gray}{Gray results} use Objects365 detection pretraining~\citep{shao2019objects365}. Non-gray ECDet rows are our primary setting; gray ECDet rows are a more compute-intensive complement that consistently improves box AP.}
	\label{tab:ecdet}
	\begin{center}
		\begin{adjustbox}{width=\textwidth}
			\begin{tabu}{l | crcc | ccc ccc}
				\toprule
				\hline
				Model & \#Epochs & \#Params. & GFLOPs & Latency (ms) & AP$^{val}$ & AP$^{val}_{50}$ & AP$^{val}_{75}$ & AP$^{val}_S$ & AP$^{val}_M$ & AP$^{val}_L$ \\
				\midrule
				\hline
				YOLOv10-S~\citep{wang2024yolov10} & 500 & 7M & 22 & 2.52 & 46.3 & 63.0 & 50.4 & 26.8 & 51.0 & 63.8 \\
				YOLO11-S~\citep{yolo11} & 500 & 9M & 22 & 7.05  & 46.6 & 63.4 & 50.3 & 28.7 & 51.3 & 64.1 \\
				YOLOv12-S-turbo~\citep{tian2025yolov12} & 600 & 9M & 19 & 8.14  & 47.6 & 64.5 & 51.5 & 28.3 & 52.7 & 65.9 \\
				RT-DETRv2-S~\citep{lv2024rt} & 120 & 20M & 60 & 4.61  & 48.1 & 65.1 & 52.1 & 30.2 & 51.5 & 63.9 \\
				D-FINE-S~\citep{peng2025dfine} & 124\textsuperscript{*} & 10M & 25 & 3.60 & 48.5 & 65.6 & 52.6 & 29.1 & 52.2 & 65.4 \\
				{DEIM}-S~\citep{huang2025deim} & 132\textsuperscript{*} & 10M & 25 & 3.60 & 49.0 & 65.9 & 53.1 & 30.4 & 52.6 & 65.7 \\
				{DEIMv2}-S~\citep{huang2025deimv2} & 132\textsuperscript{*} & 10M & 26 & 5.78 & 50.9 & 68.4 & 55.1 & 31.3 & 55.3 & 70.2 \\
				RT-DETRv4-S~\citep{liao2025rtdetrv4} & 132\textsuperscript{*} & 10M & 25 &  3.60 & 49.7 & 66.8 & 54.1 & 30.2 & 53.6 & 66.9 \\
				\rowfont{\color{gray}}LW-DETR-S~\citep{chen2024lw} & 60 & 15M & 17 & 3.09  & 48.0 & 66.9 & 51.7 & 26.8 & 52.5 & 65.5 \\
				\rowfont{\color{gray}}D-FINE-S~\citep{peng2025dfine} & - & 10M & 25 & 3.60 & 50.7 & 67.6 & 55.1 & 32.7 & 54.6 & 66.5 \\
				\rowfont{\color{gray}}RF-DETR-S~\citep{robinson2025rf} & - & 32M & 60 & 3.65  & 52.9 & 71.9 & 57.0 & 32.0 & 58.3 & 73.0 \\
				\rowfont{\color{gray}}YOLO26-S~\citep{yolo26} & 70 & 10M & 21 & 2.59 & 47.8 & 64.6 & 52.1 & 29.1 & 52.5 & 64.3  \\
                \rowfont{\color{gray}}\rowcolor[gray]{0.9} \textbf{{\detector}-S} & 74 & 10M & 26 & 5.41 & \textbf{53.6} & \textbf{71.0} & \textbf{58.0} & \textbf{35.1} & \textbf{58.3} & \textbf{71.8} \\
                
				\rowcolor[gray]{0.9} \textbf{{\detector}-S} & 74 & 10M & 26 & 5.41 & \textbf{51.7} & \textbf{69.4} & \textbf{55.8} & \textbf{32.3} & \textbf{56.4} & \textbf{70.5} \\

				\midrule
				YOLOv10-M~\citep{wang2024yolov10} & 500 & 15M & 59 & 4.70 & 51.1 & 68.1 & 55.8 & 33.8 & 56.5 & 67.0 \\
				YOLO11-M~\citep{yolo11} & 500 & 20M & 68 & 9.02 & 51.2 & 67.9 & 55.3 & 33.0 & 56.7 & 67.5 \\
				YOLOv12-M-turbo~\citep{tian2025yolov12} & 600 & 20M & 60 & 10.75 & 52.5 & 69.9 & 57.1 & 35.2 & 57.8 & 69.7 \\
				RT-DETRv2-M~\citep{lv2024rt} & 120 & 31M & 92 & 6.91  & 49.9 & 67.5 & 54.1 & 32.0 & 53.2 & 66.5 \\
				D-FINE-M~\citep{peng2025dfine} & 124\textsuperscript{*} & 19M & 57 & 5.66  & 52.3 & 69.8 & 56.4 & 33.2 & 56.5 & 70.2 \\
				DEIM-M~\citep{huang2025deim} & 102\textsuperscript{*} & 19M & 57 & 5.66 & 52.7 & 70.0 & 57.3 & 35.3 & 56.7 & 69.5 \\
				{DEIMv2}-M~\citep{huang2025deimv2} & 102\textsuperscript{*} & 18M & 52 & 8.80 & 53.0 & 70.2 & 57.6 & 34.2 & 57.4 & 71.5 \\
				RT-DETRv4-M~\citep{liao2025rtdetrv4} &  102\textsuperscript{*}& 19M & 57 & 5.66  & 53.5 & 71.1 & 58.1 & 34.9 & 57.7 & 72.1 \\
				\rowfont{\color{gray}}LW-DETR-M~\citep{chen2024lw} & 60 & 28M & 43 & 5.27  & 52.6 & 69.9 & 56.7 & 32.6 & 57.7 & 70.7 \\
				\rowfont{\color{gray}}D-FINE-M~\citep{peng2025dfine} & - & 19M & 57 & 5.66  & 55.1 & 72.6 & 59.7 & 37.9 & 59.4 & 71.7 \\
				\rowfont{\color{gray}} RF-DETR-M~\citep{robinson2025rf} & - & 34M & 79 & 4.62  & 54.7 & 73.5 & 59.2 & 36.1 & 59.7 & 73.8 \\
				\rowfont{\color{gray}}YOLO26-M~\citep{yolo26} & 80 & 20M & 68 & 4.54 & 52.5 & 69.8 & 57.2 & 36.2 & 56.9 & 68.5 \\

                \rowfont{\color{gray}}\rowcolor[gray]{0.9} \textbf{{\detector}-M} & 62 & 19M & 53 & 7.98 & \textbf{56.7} & \textbf{74.3} & \textbf{61.8} & \textbf{38.9} & \textbf{61.5} & \textbf{74.0} \\
				\rowcolor[gray]{0.9} \textbf{{\detector}-M} & 62 & 19M & 53 & 7.98 & \textbf{54.3} & \textbf{72.2} & \textbf{58.7} & \textbf{35.9} & \textbf{59.1} & \textbf{72.7} \\

				\midrule
				YOLOv10-L~\citep{wang2024yolov10} & 500 & 24M & 120 & 7.38 & 53.2 & 70.1 & 58.1 & 35.8 & 58.5 & 69.4 \\
				YOLO11-L~\citep{yolo11} & 500 & 25M & 87 & 10.34 & 53.4 & 70.1 & 58.2 & 35.6 & 59.1 & 69.2 \\
				YOLOv12-L-turbo~\citep{tian2025yolov12} & 600 & 27M & 82 & 14.39 & 53.8 & 71.0 & 58.6 & 36.9 & 59.4 & 71.0 \\
				RT-DETRv2-L~\citep{lv2024rt} & 72 & 42M & 136 & 9.29 & 53.4 & 71.6 & 57.4 & 36.1 & 57.9 & 70.8 \\
				D-FINE-L~\citep{peng2025dfine} & 74\textsuperscript{*} & 31M & 91 & 8.10 & 54.0 & 71.6 & 58.4 & 36.5 & 58.0 & 71.9 \\
				DEIM-L~\citep{huang2025deim} & 58\textsuperscript{*} & 31M & 91 & 8.10 & 54.7 & 72.4 & 59.4 & 36.9 & 59.6 & 71.8 \\
				{DEIMv2}-L~\citep{huang2025deimv2} & 68\textsuperscript{*} & 32M & 97 & 10.47 & 56.0 & 73.5 & 61.1 & 37.6 & 60.9 & 74.9 \\
				RT-DETRv4-L~\citep{liao2025rtdetrv4}& 58\textsuperscript{*} & 31M & 91 & 8.10 & 55.4 & 73.0 & 60.3  & 37.1 & 60.1 & 72.9 \\
				\rowfont{\color{gray}}LW-DETR-L~\citep{chen2024lw} & 60 & 47M & 72 & 8.25  & 56.1 & 74.6 & 60.9 & 37.2 & 60.4 & 73.0 \\
				\rowfont{\color{gray}}D-FINE-L~\citep{peng2025dfine} & - & 31M & 91 & 8.10 & 57.1 & 74.7 & 62.0 & 40.0 & 61.5 & 74.2 \\
				\rowfont{\color{gray}}RF-DETR-L~\citep{robinson2025rf} & - & 34M & 126 & 7.38  & 56.5 & 75.1 & 61.3 & 39.0 & 61.0 & 73.9 \\
				\rowfont{\color{gray}}YOLO26-L~\citep{yolo26} & 60 & 25M & 86 & 6.20 & 54.3 & 71.5 & 59.4 & 37.8 & 58.6 & 70.3 \\

                \rowfont{\color{gray}}\rowcolor[gray]{0.9} \textbf{\detector-L} & 50 & 33M & 101 & 10.49 & \textbf{59.0} & \textbf{76.9} & \textbf{64.3} & \textbf{42.0} & \textbf{64.1} & \textbf{76.1} \\
				\rowcolor[gray]{0.9} \textbf{\detector-L} & 50 & 33M & 101 & 10.49 & \textbf{57.0} & \textbf{75.1} & \textbf{61.7} & \textbf{38.7} & \textbf{62.5} & \textbf{75.0} \\

				\midrule
				YOLOv10-X~\citep{wang2024yolov10} & 500 & 30M & 160 & 10.47 & 54.4 & 71.3 & 59.3 & 37.0 & 59.8 & 70.9 \\
				YOLO11-X~\citep{yolo11} & 500 & 57M & 195 & 15.36 & 54.7 & 71.6 & 59.5 & 37.7 & 59.7 & 70.2 \\
				YOLOv12-X-turbo~\citep{tian2025yolov12} & 600 & 59M & 185 & 21.58 & 55.4 & 72.5 & 60.3 & 38.9 & 60.8 & 70.9 \\
				RT-DETRv2-X~\citep{lv2024rt} & 72 & 76M & 259 & 13.88 & 54.3 & 72.8 & 58.8 & 35.8 & 58.8 & 72.1 \\
				D-FINE-X~\citep{peng2025dfine} & 74\textsuperscript{*} & 62M & 202 & 12.90 & 55.8 & 73.7 & 60.2 & 37.3 & 60.5 & 73.4 \\
				DEIM-X~\citep{huang2025deim} & 58\textsuperscript{*} & 62M & 202 & 12.90 & 56.5 & 74.0 & 61.5 & 38.8 & 61.4 & 74.2 \\
				{DEIMv2}-X~\citep{huang2025deimv2} & 58\textsuperscript{*} & 50M & 152 & 13.75 & 57.8 & 75.3 & 63.2 & 39.1 & 62.9 & 75.9 \\
				RT-DETRv4-X~\citep{liao2025rtdetrv4} & 58\textsuperscript{*} & 62M & 202 & 12.90  & 57.0 & 74.6 & 62.1 & 39.5 & 61.9 & 74.8 \\
				\rowfont{\color{gray}}LW-DETR-X~\citep{chen2024lw} & 60 & 118M & 174 & 16.06  & 58.3 & 76.9 & 63.3 & 40.9 & 63.3 & 74.8 \\
				\rowfont{\color{gray}}D-FINE-X~\citep{peng2025dfine} & - & 62M & 202 & 12.90 & 59.3 & 76.8 & 64.6 & 42.3 & 64.2 & 76.4 \\
				\rowfont{\color{gray}}RF-DETR-X~\citep{robinson2025rf} & - & 126M & 300 & 14.79  & 58.6 & 77.4 & 63.8 & 40.3 & 63.9 & 76.2 \\
				\rowfont{\color{gray}}YOLO26-X~\citep{yolo26}  & 40 & 55M & 194& 11.65 & 56.9 & 74.1 & 62.1 & 41.3 & 61.2 & 72.7 \\

                \rowfont{\color{gray}}\rowcolor[gray]{0.9} \textbf{\detector-X} & 50 & 49M & 151 & 12.70 & \textbf{59.9} & \textbf{77.6} & \textbf{65.5} & \textbf{42.2} & \textbf{65.1} & \textbf{76.6}  \\
				\rowcolor[gray]{0.9} \textbf{\detector-X} & 50 & 49M & 151 & 12.70 & \textbf{57.9} & \textbf{76.0} & \textbf{62.9} & \textbf{38.7} & \textbf{63.4} & \textbf{76.1}  \\
				\bottomrule
			\end{tabu}
		\end{adjustbox}
	\end{center}
\end{minipage}
\clearpage
}

\paragraph{Training objective.}
ECPose follows the same set-based matching paradigm as \detector, with matching performed at the person-instance level. After matching, the training loss combines a Varifocal-style classification objective, a visible-joint regression objective, and an OKS loss:
\[
\mathcal{L}_{\mathrm{pose}} = \lambda_{\mathrm{cls}}\mathcal{L}_{\mathrm{cls}} + \lambda_{\mathrm{kpt}}\mathcal{L}_{\mathrm{kpt}} + \lambda_{\mathrm{oks}}\mathcal{L}_{\mathrm{oks}}.
\]
The coefficient $\lambda_{\mathrm{cls}}$ weights pose-quality-aware classification, while $\lambda_{\mathrm{kpt}}$ and $\lambda_{\mathrm{oks}}$ balance direct coordinate supervision and OKS-based pose quality. For the classification loss $\mathcal{L}_{\mathrm{cls}}$, we use a Varifocal Loss-style objective~\citep{zhang2021varifocalnet, janampa2025detrpose}, where the target score is defined by the standard COCO Object Keypoint Similarity (OKS)~\citep{lin2014microsoft}. Here, $\mathrm{OKS}(i,j)$ measures the similarity between prediction $i$ and ground truth $j$ by normalizing the per-keypoint localization error with the person scale and keypoint-specific constants, and averaging over visible keypoints. For the keypoint regression loss $\mathcal{L}_{\mathrm{kpt}}$, we use an $\ell_1$ loss between the predicted and ground-truth keypoint coordinates, applied only to visible joints:
\[
\mathcal{L}_{\mathrm{kpt}}=\sum_{k=1}^{K} v_{j,k}\left(|\hat{x}_{i,k}-x_{j,k}| + |\hat{y}_{i,k}-y_{j,k}|\right),
\]
where $v_{j,k}$ is the visibility indicator of the $k$-th ground-truth keypoint. Following DETRPose~\citep{janampa2025detrpose}, we also directly optimize the standard OKS loss: $\mathcal{L}_{\mathrm{oks}} = 1 - \mathrm{OKS}(i,j)$.

\subsubsection{ECInsSeg}

\paragraph{Mask head.}
We extend \detector~to instance segmentation by attaching a lightweight query-based mask head, while keeping the backbone, encoder, and decoder unchanged. The same decoder queries used for box prediction are reused for mask prediction, so the task-specific change is confined to the mask branch. Following the general design used in MaskDINO~\citep{li2023maskdino} and RF-DETR~\citep{robinson2025rf}, we build this branch from a higher-resolution encoded feature map.

Specifically, we use $\mathcal{F}^{(8)}$ as the input to the mask branch and resize it to the target mask resolution. The resized feature is processed by a depthwise convolution and a lightweight MLP to produce dense pixel embeddings, while each decoder query is projected into the same embedding space with a lightweight MLP. The mask logits for each query are then obtained by taking the dot product between its projected query embedding and the dense pixel embeddings at every spatial location. This design adds dense mask prediction with minimal extra parameters while preserving the shared ECDet representation.

\paragraph{Training objective.}
Training follows the standard DETR assignment procedure, extended with mask supervision. The overall instance segmentation loss combines the detection losses with the mask losses:
\[
\mathcal{L}_{\mathrm{insseg}} = \mathcal{L}_{\mathrm{det}} + \lambda_{\mathrm{mask}}\mathcal{L}_{\mathrm{mask}} + \lambda_{\mathrm{dice}}\mathcal{L}_{\mathrm{dice}},
\]
The coefficient $\lambda_{\mathrm{mask}}$ weights the dense mask supervision, while $\lambda_{\mathrm{dice}}$ balances region-overlap quality relative to the inherited detection objective. For the mask term $\mathcal{L}_{\mathrm{mask}}$, we use a pixel-wise mask classification loss applied only to positive matched queries. For the overlap term $\mathcal{L}_{\mathrm{dice}}$, we use a standard Dice loss~\citep{milletari2016v} on the matched masks. The Hungarian matching procedure is extended to incorporate mask supervision in addition to the standard detection costs. Because the backbone and encoder are unchanged, \seg\ provides a direct test of whether the representation distilled for detection transfers to a second dense prediction task. 
\section{Experiments}

We evaluate EdgeCrafter on three dense prediction tasks: object detection, human pose estimation, and instance segmentation. Unless otherwise specified, all models are trained and evaluated on COCO~\citep{lin2014microsoft}, and we report the standard task-specific metrics defined by the official evaluation protocol. Note that implementation details are provided in the Appendix~\ref{sec:appendix_implementationdetails}.

Hereafter, ``Extra Sup. = --'' means no task annotations beyond COCO; it does not imply the absence of upstream pretraining. All EdgeCrafter backbones inherit DINOv3 teacher supervision through feature distillation on ImageNet-1K and COCO images, while O365 variants additionally use Objects365 annotations~\citep{shao2019objects365}.

In this work, we use “edge” to refer to the overall deployment efficiency under resource-constrained edge environments, rather than simply the absolute inference latency. In practice, the suitability of a model for edge deployment is determined by multiple factors, including detection accuracy, parameter count, computational cost (FLOPs), memory and storage requirements, and real-time inference latency. These factors are also hardware- and software-dependent: for example, edge hardware and deployment stacks may be more optimized for CNN-based operators than for ViT-based architectures, making FLOPs or model size alone insufficient to predict actual runtime performance. Therefore, we evaluate EdgeCrafter from an overall accuracy–efficiency perspective, considering both model complexity and hardware-measured latency.

\subsection{Main Results}

\paragraph{Object detection.}

Table~\ref{tab:ecdet} compares ECDet with recent efficient detectors on COCO and highlights a strong accuracy--efficiency trade-off across all model scales under the non-O365 task-annotation setting defined above. In the small regime, \detector-S reaches 51.7 AP with 10M parameters and 50 GFLOPs, outperforming recent detectors in the same budget range such as RT-DETRv4-S~\citep{liao2025rtdetrv4} and DEIMv2-S~\citep{huang2025deimv2}, while also exceeding several larger models.

\begin{table}[!t]
\captionsetup{hypcap=false}
\caption{\textbf{Performance comparison of real-time object detectors on COCO~\citep{lin2014microsoft} \texttt{val2017}.}
* denotes the actual training epochs, including additional epochs without augmentation.
Latency is measured on an NVIDIA T4 GPU with batch size 1 under FP16 using TensorRT (v10.6), following RT-DETR~\citep{lv2024rt} and D-FINE~\citep{peng2025dfine}.
YOLO latency includes NMS.
\textcolor{gray}{Gray results} use Objects365 detection pretraining~\citep{shao2019objects365}. Non-gray ECDet rows are our primary setting; gray ECDet rows are a more compute-intensive complement that consistently improves box AP.}
\label{tab:ecdet}
	\centering
    \vspace{6pt}
		\begin{adjustbox}{width=\textwidth}
			\begin{tabu}{l | crcc | ccc ccc}
				\toprule
				\hline
				Model & \#Epochs & \#Params. & GFLOPs & Latency (ms) & AP$^{val}$ & AP$^{val}_{50}$ & AP$^{val}_{75}$ & AP$^{val}_S$ & AP$^{val}_M$ & AP$^{val}_L$ \\
				\midrule
				\hline
				YOLOv10-S~\citep{wang2024yolov10} & 500 & 7M & 22 & 2.52 & 46.3 & 63.0 & 50.4 & 26.8 & 51.0 & 63.8 \\
				YOLO11-S~\citep{yolo11} & 500 & 9M & 22 & 7.05  & 46.6 & 63.4 & 50.3 & 28.7 & 51.3 & 64.1 \\
				YOLOv12-S-turbo~\citep{tian2025yolov12} & 600 & 9M & 19 & 8.14  & 47.6 & 64.5 & 51.5 & 28.3 & 52.7 & 65.9 \\
				RT-DETRv2-S~\citep{lv2024rt} & 120 & 20M & 60 & 4.61  & 48.1 & 65.1 & 52.1 & 30.2 & 51.5 & 63.9 \\
				D-FINE-S~\citep{peng2025dfine} & 124\textsuperscript{*} & 10M & 25 & 3.60 & 48.5 & 65.6 & 52.6 & 29.1 & 52.2 & 65.4 \\
				{DEIM}-S~\citep{huang2025deim} & 132\textsuperscript{*} & 10M & 25 & 3.60 & 49.0 & 65.9 & 53.1 & 30.4 & 52.6 & 65.7 \\
				{DEIMv2}-S~\citep{huang2025deimv2} & 132\textsuperscript{*} & 10M & 50 & 5.78 & 50.9 & 68.4 & 55.1 & 31.3 & 55.3 & 70.2 \\
				RT-DETRv4-S~\citep{liao2025rtdetrv4} & 132\textsuperscript{*} & 10M & 25 &  3.60 & 49.7 & 66.8 & 54.1 & 30.2 & 53.6 & 66.9 \\
				\rowfont{\color{gray}}LW-DETR-S~\citep{chen2024lw} & 60 & 15M & 17 & 3.09  & 48.0 & 66.9 & 51.7 & 26.8 & 52.5 & 65.5 \\
				\rowfont{\color{gray}}D-FINE-S~\citep{peng2025dfine} & - & 10M & 25 & 3.60 & 50.7 & 67.6 & 55.1 & 32.7 & 54.6 & 66.5 \\
				\rowfont{\color{gray}}RF-DETR-S~\citep{robinson2025rf} & - & 32M & 60 & 3.65  & 52.9 & 71.9 & 57.0 & 32.0 & 58.3 & 73.0 \\
				\rowfont{\color{gray}}YOLO26-S~\citep{yolo26} & 70 & 10M & 21 & 2.59 & 47.8 & 64.6 & 52.1 & 29.1 & 52.5 & 64.3  \\
                \rowfont{\color{gray}}\rowcolor[gray]{0.9} \textbf{{\detector}-S} & 74 & 10M & 50 & 5.41 & \textbf{53.6} & \textbf{71.0} & \textbf{58.0} & \textbf{35.1} & \textbf{58.3} & \textbf{71.8} \\
                
				\rowcolor[gray]{0.9} \textbf{{\detector}-S} & 74 & 10M & 50 & 5.41 & \textbf{51.7} & \textbf{69.4} & \textbf{55.8} & \textbf{32.3} & \textbf{56.4} & \textbf{70.5} \\

				\midrule
				YOLOv10-M~\citep{wang2024yolov10} & 500 & 15M & 59 & 4.70 & 51.1 & 68.1 & 55.8 & 33.8 & 56.5 & 67.0 \\
				YOLO11-M~\citep{yolo11} & 500 & 20M & 68 & 9.02 & 51.2 & 67.9 & 55.3 & 33.0 & 56.7 & 67.5 \\
				YOLOv12-M-turbo~\citep{tian2025yolov12} & 600 & 20M & 60 & 10.75 & 52.5 & 69.9 & 57.1 & 35.2 & 57.8 & 69.7 \\
				RT-DETRv2-M~\citep{lv2024rt} & 120 & 31M & 92 & 6.91  & 49.9 & 67.5 & 54.1 & 32.0 & 53.2 & 66.5 \\
				D-FINE-M~\citep{peng2025dfine} & 124\textsuperscript{*} & 19M & 57 & 5.66  & 52.3 & 69.8 & 56.4 & 33.2 & 56.5 & 70.2 \\
				DEIM-M~\citep{huang2025deim} & 102\textsuperscript{*} & 19M & 57 & 5.66 & 52.7 & 70.0 & 57.3 & 35.3 & 56.7 & 69.5 \\
				{DEIMv2}-M~\citep{huang2025deimv2} & 102\textsuperscript{*} & 18M & 84 & 8.80 & 53.0 & 70.2 & 57.6 & 34.2 & 57.4 & 71.5 \\
				RT-DETRv4-M~\citep{liao2025rtdetrv4} &  102\textsuperscript{*}& 19M & 57 & 5.66  & 53.5 & 71.1 & 58.1 & 34.9 & 57.7 & 72.1 \\
				\rowfont{\color{gray}}LW-DETR-M~\citep{chen2024lw} & 60 & 28M & 43 & 5.27  & 52.6 & 69.9 & 56.7 & 32.6 & 57.7 & 70.7 \\
				\rowfont{\color{gray}}D-FINE-M~\citep{peng2025dfine} & - & 19M & 57 & 5.66  & 55.1 & 72.6 & 59.7 & 37.9 & 59.4 & 71.7 \\
				\rowfont{\color{gray}} RF-DETR-M~\citep{robinson2025rf} & - & 34M & 79 & 4.62  & 54.7 & 73.5 & 59.2 & 36.1 & 59.7 & 73.8 \\
				\rowfont{\color{gray}}YOLO26-M~\citep{yolo26} & 80 & 20M & 68 & 4.54 & 52.5 & 69.8 & 57.2 & 36.2 & 56.9 & 68.5 \\

                \rowfont{\color{gray}}\rowcolor[gray]{0.9} \textbf{{\detector}-M} & 62 & 19M & 85 & 7.98 & \textbf{56.7} & \textbf{74.3} & \textbf{61.8} & \textbf{38.9} & \textbf{61.5} & \textbf{74.0} \\
				\rowcolor[gray]{0.9} \textbf{{\detector}-M} & 62 & 19M & 85 & 7.98 & \textbf{54.3} & \textbf{72.2} & \textbf{58.7} & \textbf{35.9} & \textbf{59.1} & \textbf{72.7} \\

				\midrule
				YOLOv10-L~\citep{wang2024yolov10} & 500 & 24M & 120 & 7.38 & 53.2 & 70.1 & 58.1 & 35.8 & 58.5 & 69.4 \\
				YOLO11-L~\citep{yolo11} & 500 & 25M & 87 & 10.34 & 53.4 & 70.1 & 58.2 & 35.6 & 59.1 & 69.2 \\
				YOLOv12-L-turbo~\citep{tian2025yolov12} & 600 & 27M & 82 & 14.39 & 53.8 & 71.0 & 58.6 & 36.9 & 59.4 & 71.0 \\
				RT-DETRv2-L~\citep{lv2024rt} & 72 & 42M & 136 & 9.29 & 53.4 & 71.6 & 57.4 & 36.1 & 57.9 & 70.8 \\
				D-FINE-L~\citep{peng2025dfine} & 74\textsuperscript{*} & 31M & 91 & 8.10 & 54.0 & 71.6 & 58.4 & 36.5 & 58.0 & 71.9 \\
				DEIM-L~\citep{huang2025deim} & 58\textsuperscript{*} & 31M & 91 & 8.10 & 54.7 & 72.4 & 59.4 & 36.9 & 59.6 & 71.8 \\
				{DEIMv2}-L~\citep{huang2025deimv2} & 68\textsuperscript{*} & 32M & 144 & 10.47 & 56.0 & 73.5 & 61.1 & 37.6 & 60.9 & 74.9 \\
				RT-DETRv4-L~\citep{liao2025rtdetrv4}& 58\textsuperscript{*} & 31M & 91 & 8.10 & 55.4 & 73.0 & 60.3  & 37.1 & 60.1 & 72.9 \\
				\rowfont{\color{gray}}LW-DETR-L~\citep{chen2024lw} & 60 & 47M & 72 & 8.25  & 56.1 & 74.6 & 60.9 & 37.2 & 60.4 & 73.0 \\
				\rowfont{\color{gray}}D-FINE-L~\citep{peng2025dfine} & - & 31M & 91 & 8.10 & 57.1 & 74.7 & 62.0 & 40.0 & 61.5 & 74.2 \\
				\rowfont{\color{gray}}RF-DETR-L~\citep{robinson2025rf} & - & 34M & 126 & 7.38  & 56.5 & 75.1 & 61.3 & 39.0 & 61.0 & 73.9 \\
				\rowfont{\color{gray}}YOLO26-L~\citep{yolo26} & 60 & 25M & 86 & 6.20 & 54.3 & 71.5 & 59.4 & 37.8 & 58.6 & 70.3 \\

                \rowfont{\color{gray}}\rowcolor[gray]{0.9} \textbf{\detector-L} & 50 & 33M & 147 & 10.49 & \textbf{59.0} & \textbf{76.9} & \textbf{64.3} & \textbf{42.0} & \textbf{64.1} & \textbf{76.1} \\
				\rowcolor[gray]{0.9} \textbf{\detector-L} & 50 & 33M & 147 & 10.49 & \textbf{57.0} & \textbf{75.1} & \textbf{61.7} & \textbf{38.7} & \textbf{62.5} & \textbf{75.0} \\

				\midrule
				YOLOv10-X~\citep{wang2024yolov10} & 500 & 30M & 160 & 10.47 & 54.4 & 71.3 & 59.3 & 37.0 & 59.8 & 70.9 \\
				YOLO11-X~\citep{yolo11} & 500 & 57M & 195 & 15.36 & 54.7 & 71.6 & 59.5 & 37.7 & 59.7 & 70.2 \\
				YOLOv12-X-turbo~\citep{tian2025yolov12} & 600 & 59M & 185 & 21.58 & 55.4 & 72.5 & 60.3 & 38.9 & 60.8 & 70.9 \\
				RT-DETRv2-X~\citep{lv2024rt} & 72 & 76M & 259 & 13.88 & 54.3 & 72.8 & 58.8 & 35.8 & 58.8 & 72.1 \\
				D-FINE-X~\citep{peng2025dfine} & 74\textsuperscript{*} & 62M & 202 & 12.90 & 55.8 & 73.7 & 60.2 & 37.3 & 60.5 & 73.4 \\
				DEIM-X~\citep{huang2025deim} & 58\textsuperscript{*} & 62M & 202 & 12.90 & 56.5 & 74.0 & 61.5 & 38.8 & 61.4 & 74.2 \\
				{DEIMv2}-X~\citep{huang2025deimv2} & 58\textsuperscript{*} & 50M & 200 & 13.75 & 57.8 & 75.3 & 63.2 & 39.1 & 62.9 & 75.9 \\
				RT-DETRv4-X~\citep{liao2025rtdetrv4} & 58\textsuperscript{*} & 62M & 202 & 12.90  & 57.0 & 74.6 & 62.1 & 39.5 & 61.9 & 74.8 \\
				\rowfont{\color{gray}}LW-DETR-X~\citep{chen2024lw} & 60 & 118M & 174 & 16.06  & 58.3 & 76.9 & 63.3 & 40.9 & 63.3 & 74.8 \\
				\rowfont{\color{gray}}D-FINE-X~\citep{peng2025dfine} & - & 62M & 202 & 12.90 & 59.3 & 76.8 & 64.6 & 42.3 & 64.2 & 76.4 \\
				\rowfont{\color{gray}}RF-DETR-X~\citep{robinson2025rf} & - & 126M & 300 & 14.79  & 58.6 & 77.4 & 63.8 & 40.3 & 63.9 & 76.2 \\
				\rowfont{\color{gray}}YOLO26-X~\citep{yolo26}  & 40 & 55M & 194& 11.65 & 56.9 & 74.1 & 62.1 & 41.3 & 61.2 & 72.7 \\

                \rowfont{\color{gray}}\rowcolor[gray]{0.9} \textbf{\detector-X} & 50 & 49M & 204 & 12.70 & \textbf{59.9} & \textbf{77.5} & \textbf{65.2} & \textbf{42.4} & \textbf{64.8} & \textbf{77.0}  \\
				\rowcolor[gray]{0.9} \textbf{\detector-X} & 50 & 49M & 204 & 12.70 & \textbf{57.9} & \textbf{76.0} & \textbf{62.9} & \textbf{38.7} & \textbf{63.4} & \textbf{76.1}  \\
				\bottomrule
			\end{tabu}
		\end{adjustbox}
\end{table}

The same trend continues as the model scales up: \detector-M reaches 54.3 AP with 19M parameters, \detector-L improves to 57.0 AP with 33M parameters, and \detector-X reaches 57.9 AP with 49M parameters and 204 GFLOPs. Notably, these larger variants remain competitive with, and in several cases outperform, baselines that benefit from additional Objects365 pretraining~\citep{shao2019objects365}, indicating that task-specialized distillation can partially offset the need for an additional labeled task dataset such as Objects365.

In terms of latency, ECDet is not always the absolute fastest model in every budget range, which is expected given the current software and hardware optimization gap for ViT-based architectures. However, the measured runtimes remain within a practical range for real-time or near-real-time deployment on the reported hardware, while the accuracy gains are substantial. This trade-off is still meaningful for real applications, since edge deployment is often constrained not only by runtime but also by available compute and storage budgets. For example, commercial SoCs such as Rockchip RK3568 integrate only a 1 TOPS NPU for INT8 inference, making compact models attractive not only for throughput but also for reducing flash and memory usage~\citep{rockchip_rk3568}. Overall, the results show that the distilled compact ViT backbone scales favorably for edge object detection without additional task-annotation datasets beyond COCO in the non-O365 setting or substantially larger detectors.

\begin{table}[t]
	\caption{\textbf{Performance comparison of real-time pose estimation methods on COCO \texttt{val2017}, grouped by model scale.} The \textbf{Extra Sup.} column reports additional task annotations beyond COCO; \texttt{O365} denotes Objects365 detection pretraining~\citep{shao2019objects365}. Non-O365 ECPose rows are our primary setting; O365 rows are a more compute-intensive complement that consistently improves keypoint AP.}
	\label{tab:pose_res}
	\begin{center}
		\begin{adjustbox}{width=\textwidth}
			\begin{tabu}{l c | r c c | ccc cc c}
				\toprule 
				\hline
				Model & Extra Sup. & \#Params. & GFLOPs & Latency (ms) & AP$^{val}$ & AP$^{val}_{50}$ & AP$^{val}_{75}$ & AP$^{val}_{M}$ & AP$^{val}_{L}$ & AR$^{val}$\\
				\midrule
				\hline
				RTMO-S~\citep{rtmo} & -- & 9.9M & 30.7 & 3.78 & 67.7 & 87.8 & 73.7 & - & - & 71.5\\
				YOLO11-Pose-S~\citep{yolo11} & -- & 9.9M & 23.2 & 4.54  & 58.9 & 86.3 & 64.8 & 54.0 & 68.0 & 66.1 \\
                YOLO26-Pose-S~\citep{yolo26} & -- & 10.4M & 23.9 & 2.70 & 63.1 & 86.6 & 68.8 & 56.5 & 73.7 & 69.0 \\
				\rowfont{\color{gray}}DETRPose-S~\citep{janampa2025detrpose} & \textcolor{gray}{O365} & 11.5M & 33.1 & 5.12 & 67.0 & 87.6 & 72.8 & 60.2 & 77.4 & {73.5}\\

                \rowfont{\color{gray}}\rowcolor[gray]{0.9} \textbf{\pose-S} & O365 & 9.9M  & 53.1 & 5.54 & \textbf{69.7}  & \textbf{89.4}  & \textbf{76.2}  & \textbf{61.8}  & \textbf{81.5}  & \textbf{75.3}  \\
				\rowcolor[gray]{0.9} \textbf{\pose-S} & \textbf{--} & 9.9M  & 53.1 & 5.54 & \textbf{68.9}  & \textbf{89.1}  & \textbf{75.2}  & \textbf{60.7}  & \textbf{81.1}  & \textbf{74.6}  \\

				\midrule
				RTMO-M~\citep{rtmo} & -- & 22.6M & 69.0 & 6.21 & 70.9 & 89.0 & 77.8 & - & - & 74.7 \\
				YOLO11-Pose-M~\citep{yolo11} & -- & 20.9M & 71.7 & 6.65 & 64.9 & 89.4 & 72.4 & 62.2 & 71.6 & 72.2 \\
                YOLO26-Pose-M~\citep{yolo26} & -- & 21.5M & 73.1 & 4.75 & 68.8 & 89.6 & 75.5 & 64.0 & 77.2 & 74.6 \\
				\rowfont{\color{gray}}DETRPose-M~\citep{janampa2025detrpose} & \textcolor{gray}{O365} & 20.8M & 67.3 & 7.90 & 69.4 & 89.2 & 75.4 & 63.2 & 79.0 & 75.5\\

                \rowfont{\color{gray}}\rowcolor[gray]{0.9} \textbf{\pose-M} & O365 & 19.8M & 92.1 & 9.25 & \textbf{73.1} & \textbf{91.2} & \textbf{79.4} & \textbf{65.7} & \textbf{84.4} & \textbf{78.7} \\
				\rowcolor[gray]{0.9} \textbf{\pose-M} & \textbf{--} & 19.8M & 92.1 & 9.25 & \textbf{72.4} & \textbf{90.9} & \textbf{78.6} & \textbf{65.2} & \textbf{83.6} & \textbf{78.2} \\
				
				\midrule
				RTMO-L~\citep{rtmo} & -- & 44.8M & 136.7 & 8.35 & {72.4} & 89.9 & {78.8} & - & - & 76.8\\
				YOLO11-Pose-L~\citep{yolo11} & -- & 26.2M & 90.7 & 7.95 & 66.1 & 89.9 & 73.6 & 63.2 & 73.1 & 73.3 \\
                YOLO26-Pose-L~\citep{yolo26} & -- & 25.9M & 91.3 & 6.18 & 70.4 & 90.5 & 77.4 & 65.7 & 78.4 & 75.9 \\
				\rowfont{\color{gray}}DETRPose-L~\citep{janampa2025detrpose} & \textcolor{gray}{O365} & {32.8M} & 107.1 & 11.51 & 72.5 & {90.6} & 79.0 & 66.3 & {82.2} & {78.7}\\

                \rowfont{\color{gray}}\rowcolor[gray]{0.9} \textbf{\pose-L} & O365 & 34.3M  & 156.7 & 11.83 & \textbf{74.5}  & \textbf{92.2}  & \textbf{81.1}  & \textbf{67.6}  & \textbf{85.0}  & \textbf{79.9}  \\ 
				\rowcolor[gray]{0.9} \textbf{\pose-L} & \textbf{--} & 34.3M  & 156.7 & 11.83 & \textbf{73.5}  & \textbf{91.7}  & \textbf{79.9}  & \textbf{66.4}  & \textbf{84.4}  & \textbf{78.8}  \\ 
				
				\midrule
				ED-Pose~\citep{edpose} & -- & 218.0M & 422.6 & - & 74.3 & 91.5 & 81.7 & 68.5 & 82.7 & -\\
				YOLO11-Pose-X~\citep{yolo11} & -- & 58.8M & 203.3 & 12.95 & 69.5 & 91.1 & 77.4 & 66.6 & 76.0 & 76.3 \\
                YOLO26-Pose-X~\citep{yolo26} & -- & 57.6M & 201.7 & 11.05 & 71.6 & 91.6 & 78.9 & 67.4 & 79.5 & 77.2 \\
				\rowfont{\color{gray}}DETRPose-X~\citep{janampa2025detrpose} & \textcolor{gray}{O365} & 73.3M & 239.5 & 18.89 & 73.3 & 90.5 & 79.4 & 67.5 & 82.7 & 79.4\\
                \rowfont{\color{gray}}\rowcolor[gray]{0.9} \textbf{\pose-X} & O365 & 50.6M & 217.2 & 14.31 & \textbf{75.9} & \textbf{92.4} & \textbf{82.7} & \textbf{69.5} & \textbf{85.9} & \textbf{81.2} \\
				\rowcolor[gray]{0.9} \textbf{\pose-X} & \textbf{--} & 50.6M & 217.2 & 14.31 & \textbf{74.8} & \textbf{92.2} & \textbf{81.5} & \textbf{68.0} & \textbf{85.4} & \textbf{80.1} \\
				\bottomrule
			\end{tabu}
		\end{adjustbox}
	\end{center}
\end{table}

\paragraph{Transfer to human pose estimation.}

Table \ref{tab:pose_res} shows that the detection-distilled representation transfers effectively to human pose estimation with task-specific decoder changes and pose-training annotations restricted to COCO. In the small-scale regime, \pose-S reaches 68.9 AP with only 9.9M parameters, significantly outperforming YOLO11-Pose-S~\citep{yolo11} (58.9 AP) and YOLO26-Pose-S~\citep{yolo26} (63.1 AP), while also exceeding the Objects365-pretrained~\citep{shao2019objects365} DETRPose-S~\citep{janampa2025detrpose} (67.0 AP). This trend continues at medium scale, where \pose-M achieves 72.4 AP, not only surpassing its direct competitors but also matching the performance of much larger models such as RTMO-L~\citep{rtmo} and DETRPose-L. At larger scales, \pose-L and \pose-X reach 73.5 AP and 74.8 AP, respectively, consistently outperforming the corresponding DETRPose models. Notably, \pose-X even surpasses the heavy-weight ED-Pose~\citep{edpose} (74.3 AP) despite the latter having a massive 218M parameter count. The latency numbers again show that the proposed models are practical, even if they are not always the lowest-latency option in every scale regime, and the improved accuracy is obtained without expanding to substantially larger backbones. Taken together, these results indicate that the representation learned through detection-centered distillation is not limited to bounding-box prediction: it also transfers effectively to fine-grained keypoint localization while preserving a strong accuracy--efficiency trade-off across the full model family.

\begin{table}[t]
	\caption{\textbf{Performance comparison of real-time instance segmentation models on COCO~\citep{lin2014microsoft} \texttt{val2017}, grouped by model scale.} 
		 The \textbf{Extra Sup.} column reports additional task annotations beyond COCO. \texttt{O365} denotes Objects365 detection pretraining~\citep{shao2019objects365} without pseudo masks; \texttt{O365+SAM2} additionally uses SAM2~\citep{ravi2025sam} pseudo-instance masks. Non-O365 ECInsSeg rows are our primary setting; O365 rows are a more compute-intensive complement that consistently improves mask AP.} 
	\label{tab:XLMS-segm}
	\begin{center}
		\begin{adjustbox}{width=\textwidth}
			\begin{tabu}{l c | rcc |c ccc ccc}
				\toprule
				\hline
				Model 
				& Extra Sup.
				& \#Params. 
				& GFLOPs 
				& Latency (ms) 
				& AP$^{val}$ 
				& AP$^{val}_{50}$ 
				& AP$^{val}_{75}$ 
				& AP$^{val}_S$ 
				& AP$^{val}_M$ 
				& AP$^{val}_L$ \\
				\midrule
				\hline
				YOLOv8-Seg-S~\citep{yolov8}  & -- & 11.8M & 42.6 & 7.08  & 36.8 & 58.1 & 38.9 & 17.3 & 41.3 & 53.7 \\
				YOLO11-Seg-S~\citep{yolo11} & -- & 10.1M & 35.5 & 7.20  &  37.8 & 59.9 & 40.4 & 19.8 & 42.7 & 55.2 \\
				\rowfont{\color{gray}}YOLO26-Seg-S~\citep{yolo26} & \textcolor{gray}{O365+SAM2} & 10.4M  & 34.2 & 3.51 & 40.0 & 61.5 & 43.0 & 21.0 & 44.5 & 57.3 \\
				\rowfont{\color{gray}}RF-DETR-Seg-S~\citep{robinson2025rf} & \textcolor{gray}{O365+SAM2} & 33.7M  & 70.6 & 4.81  & 43.1 & 66.2 & 45.9 & 21.9 & 48.5 & 64.1 \\

                \rowfont{\color{gray}}\rowcolor[gray]{0.9} \textbf{\seg-S} & O365 & 10.3M & 57.0 & 6.96 & \textbf{43.9} & \textbf{66.7} & \textbf{46.9} & \textbf{21.6} & \textbf{47.1} & \textbf{66.7} \\
				\rowcolor[gray]{0.9} \textbf{\seg-S} & \textbf{--} & 10.3M & 57.0 & 6.96 & \textbf{43.0} & \textbf{65.7} & \textbf{46.0} & \textbf{20.8} & \textbf{46.3} & \textbf{65.9} \\

				\hline
				YOLOv8-Seg-M~\citep{yolov8}  & -- & 27.3M & 110.2 & 9.61  & 40.8 & 63.6 & 43.8 & 22.0 & 46.0 & 58.2 \\
				YOLO11-Seg-M~\citep{yolo11} & -- & 22.4M & 113.2 & 9.18  & 41.5 & 65.1 & 44.5 & \textbf{23.0} & 47.2 & 59.1 \\
				\rowfont{\color{gray}}YOLO26-Seg-M~\citep{yolo26} & \textcolor{gray}{O365+SAM2} & 23.6M  & 121.5 & 6.53  & 44.1& 66.8 & 47.7 & 25.6 & 48.9 & 60.2 \\
				\rowfont{\color{gray}} RF-DETR-Seg-M~\citep{robinson2025rf} & \textcolor{gray}{O365+SAM2} & 35.7M  & 102.0 & 6.35 & 45.3 & 68.4 & 48.8 & 25.5 & 50.4 & 65.3 \\
                
                \rowfont{\color{gray}}\rowcolor[gray]{0.9} \textbf{\seg-M} & O365 & 20.1M & 96.1  & 9.85 & \textbf{46.9} & \textbf{70.2} & \textbf{50.5} & 25.2 & \textbf{50.7} & \textbf{69.0} \\
				\rowcolor[gray]{0.9} \textbf{\seg-M} & \textbf{--} & 20.1M & 96.1  & 9.85 & \textbf{45.2} & \textbf{68.2} & \textbf{48.3} & 22.9 & \textbf{49.0} & \textbf{68.1} \\

				\hline
				YOLOv8-Seg-L~\citep{yolov8}  & -- & 46.0M & 220.5 & 12.29  & 42.6 & 66.2 & 45.5 & 23.9 & 47.9 & 59.5 \\
				YOLO11-Seg-L~\citep{yolo11} & -- & 27.6M & 132.2 & 10.55  & 42.9 & 67.0 & 46.2 & 24.6 & 48.7 & 60.6 \\
				\rowfont{\color{gray}}YOLO26-Seg-L~\citep{yolo26} & \textcolor{gray}{O365+SAM2} & 28.0M  & 139.8 & 7.77  & 45.5 & 68.7 & 49.2 & 27.1 & 50.4 & 62.8     \\        
				\rowfont{\color{gray}}RF-DETR-Seg-L~\citep{robinson2025rf} & \textcolor{gray}{O365+SAM2} & 36.2M  & 151.1 & 9.42  & 47.1 & 70.5 & 50.9 & 28.4 & 52.1 & 65.6 \\   

                \rowfont{\color{gray}}\rowcolor[gray]{0.9} \textbf{\seg-L}   & O365 & 33.6M & 158.3 & 12.56 & \textbf{48.8} & \textbf{72.9} & \textbf{53.0} & \textbf{27.4} & \textbf{52.8} & \textbf{71.0} \\
				\rowcolor[gray]{0.9} \textbf{\seg-L}   & \textbf{--} & 33.6M & 158.3 & 12.56 & \textbf{47.1} & \textbf{70.9} & \textbf{50.5} & \textbf{24.8} & \textbf{51.1} & \textbf{69.6} \\

				\hline
				MaskDINO~\citep{li2023maskdino} & -- & 52.1M & 286 & -  & 46.3 & 69.0 & 50.7 & 26.1 & 49.3 & 66.1 \\
				YOLOv8-Seg-X~\citep{yolov8}  & -- & 71.8M & 344.1 & 16.20  & 43.4 & 67.1 & 46.5 & 25.6 & 48.9 & 60.4 \\
				YOLO11-Seg-X~\citep{yolo11} & -- & 62.1M & 296.4 & 15.54  & 43.8 & 68.5 & 47.1 & 25.7 & 49.7 & 61.4 \\
				\rowfont{\color{gray}}YOLO26-Seg-X~\citep{yolo26} & \textcolor{gray}{O365+SAM2} & 62.8M  & 313.5 & 15.13  & 47.0 & 70.8 & 51.1 & 29.7 & 51.8 & 63.1 \\
				\rowfont{\color{gray}}RF-DETR-Seg-X~\citep{robinson2025rf} & \textcolor{gray}{O365+SAM2} & 38.1M  & 260.0 & 15.42 & 48.8 & 72.2 & 53.1 & 30.6 & 53.3 & 65.9 \\

                \rowfont{\color{gray}}\rowcolor[gray]{0.9} \textbf{\seg-X}  & O365 & 49.9M & 215.6 & 14.96 & \textbf{49.8} & \textbf{73.7} & \textbf{53.8} & \textbf{27.6} & \textbf{53.9} & \textbf{71.9} \\
				\rowcolor[gray]{0.9} \textbf{\seg-X}  & \textbf{--} & 49.9M & 215.6 & 14.96 & \textbf{48.4} & \textbf{72.2} & \textbf{52.0} & \textbf{26.3} & \textbf{52.7} & \textbf{71.1} \\
				\bottomrule
			\end{tabu}
		\end{adjustbox}
	\end{center}
\end{table}

\paragraph{Transfer to instance segmentation.}

Table~\ref{tab:XLMS-segm} evaluates whether the same detection-distilled representation can support dense mask prediction with a lightweight segmentation head whose task-specific training uses only COCO instance annotations. In the small regime, \seg-S reaches 43.0 AP with only 10.3M parameters, clearly outperforming YOLO11-Seg-S~\citep{yolo11} and nearly matching RF-DETR-Seg-S~\citep{robinson2025rf} even though the latter is much larger and uses additional Objects365 pretraining~\citep{shao2019objects365} together with pseudo masks from SAM2~\citep{ravi2025sam}. The same behavior remains stable at higher capacities: \seg-M reaches 45.2 AP and stays very close to RF-DETR-Seg-M with a smaller model, while \seg-L matches RF-DETR-Seg-L at 47.1 AP with a comparable FLOP budget. At the extra-large scale, \seg-X reaches 48.4 AP with 215.6 GFLOPs, approaching RF-DETR-Seg-X while remaining substantially more compute-efficient. Similar to detection and pose, the reported latency is practical for deployment even though ECInsSeg is not always the fastest entry in the table, and the compact parameter count remains important for memory- and storage-constrained devices. Overall, these results show that the distilled ECDet representation transfers beyond detection to instance segmentation without requiring a separate backbone redesign, while preserving a favorable mask-accuracy, parameter, and FLOP trade-off across the full scale range.

\paragraph{Complementary Objects365 Pretraining.}
Our primary non-O365 models already perform strongly without the additional compute of Objects365 pretraining~\citep{shao2019objects365}. As a complementary, more compute-intensive setting, O365 consistently improves every scale; the X variants reach 59.9 box AP, 49.8 mask AP, and 75.9 keypoint AP. The pose and segmentation gains transfer from the O365-pretrained detector without Objects365 keypoint or mask annotations or pseudo-mask supervision.

\subsection{Analysis}

\subsubsection{Distillation Analysis}

We first analyze the distillation recipe, since the paper's main claim is that compact ViTs become competitive when the teacher, data, and optimization strategy are chosen to match dense prediction. Unless otherwise specified, all ablations distill ECViT-T+ and evaluate the resulting backbone through the downstream performance of ECDet-M on COCO~\citep{lin2014microsoft}. We use downstream detection AP rather than the distillation loss as the evaluation metric, because lower feature-matching loss does not always translate into better dense-prediction performance.

\begin{table}[!t]
\centering
\caption{\textbf{Unified ablation on teacher design and distillation data.} All experiments use ECViT-T+ as the student. The teacher is initialized from DINOv3~\citep{simeoni2025dinov3}, and ``COCO pre-training'' indicates that the teacher is first adapted as the ECDet backbone on COCO before distillation.}
\label{tab:ablation_unified}

\begin{tabular}{l c l c}
\toprule
Teacher Arch. & COCO Pre-training & Distillation Dataset & AP (\%) \\
\midrule
\multicolumn{4}{l}{\textit{Teacher capacity}} \\
DINOv3-S & \cmark & IN-1K + COCO & 54.0 \\
\rowcolor[gray]{0.95}\textbf{DINOv3-B} & \cmark & IN-1K + COCO & \textbf{54.3} \\
DINOv3-L & \cmark & IN-1K + COCO & 52.6 \\
\midrule
\multicolumn{4}{l}{\textit{Teacher task adaptation}} \\
DINOv3-B & \xmark & IN-1K + COCO & 53.5 \\
\midrule
\multicolumn{4}{l}{\textit{Distillation dataset}} \\
DINOv3-B & \cmark & IN-1K & 54.1 \\
\bottomrule
\end{tabular}
\end{table}

\paragraph{Teacher design and distillation data.}
Table~\ref{tab:ablation_unified} summarizes the most important design choices in the distillation pipeline. First, teacher capacity should be matched to the student rather than increased without bound: DINOv3-B~\citep{simeoni2025dinov3} gives the best result, while the smaller DINOv3-S is weaker and the larger DINOv3-L degrades performance despite its higher capacity. This suggests that an excessively strong teacher can create a representation gap that the compact student cannot effectively absorb. Second, adapting the teacher to detection on COCO before distillation improves AP from 53.5 to 54.3, confirming that teacher specificity matters for dense prediction. Third, adding COCO images to the distillation corpus improves AP from 54.1 to 54.3 over using ImageNet-1K alone~\citep{deng2009imagenet}, indicating that additional task-relevant images might provide complementary supervision beyond generic classification data. Taken together, these results support the final recipe used in EdgeCrafter: a detection-specialized, scale-matched teacher distilled on both ImageNet-1K and COCO.

\begin{table}[!t]
\centering
\begin{minipage}[t]{0.5\linewidth}
\vspace{0pt}
\centering
\caption{\textbf{Ablation on feature-alignment depth.} We vary how many of the teacher's final layers supervise the student and report downstream detection AP after distillation.}

\label{tab:ablation_layers}
\tabcolsep=0.098cm
\renewcommand{\arraystretch}{0.8}
\begin{tabular}{lccc}
\toprule
Student & Student Layers & Teacher Layers & AP \\
\midrule
\multirow{5}{*}{ECViT-T+} & 1 & 1 & 53.6 \\
 & 1 & 2 & 54.3 \\
 & 1 & 3 & \textbf{54.6} \\
 & 1 & 4 & 54.1 \\
 & 2 & 2 & 54.2 \\
\midrule
\multirow{2}{*}{ECViT-S} & 1 & 2 & \textbf{57.0} \\
 & 1 & 3 & 56.9 \\
\midrule
\multirow{2}{*}{ECViT-S+} & 1 & 2 & \textbf{57.9} \\
 & 1 & 3 & 57.5 \\
\bottomrule
\end{tabular}
\end{minipage}
\hfill
\begin{minipage}[t]{0.45\linewidth}
\centering
\caption{\textbf{Ablation on optimization and register tokens.} Unless otherwise specified, the default setting uses LARS~\citep{you2017large} and one register token~\citep{darcet2023vision}.}

\label{tab:ablation_optimizer}
\renewcommand{\arraystretch}{1.05}
\begin{tabular}{ccc}
\toprule
Optimizer & Register Number & AP \\
\midrule
\multicolumn{3}{c}{\textbf{Effect of optimizer}} \\
AdamW & 1 & 54.0 \\
\rowcolor[gray]{0.95}  LARS  & 1 & \textbf{54.3} \\
\midrule
\multicolumn{3}{c}{\textbf{Effect of register tokens}} \\
LARS & 0 & 53.8 \\
LARS & 2 & 54.2 \\
LARS & 4 & 54.2 \\
\bottomrule
\end{tabular}
\end{minipage}
\end{table}

\paragraph{Feature alignment depth.}
Table~\ref{tab:ablation_layers} studies how many late teacher features should supervise the student. Pure one-to-one alignment between the last teacher and student layers is clearly suboptimal, suggesting that a single-point matching objective does not transfer enough high-level supervision. Aligning the student's final layer to the teacher's last two layers improves performance consistently across all tested scales, and this is the configuration adopted in the final model. Although using the last three teacher layers gives a slight gain for ECViT-T+, that advantage does not carry over to the larger ECViT-S and ECViT-S+ variants. We therefore choose the last-two-layer alignment because it is the most stable option across scales, even if the very best result for one particular student is obtained with three layers.

\paragraph{Optimization and register tokens.}
Table~\ref{tab:ablation_optimizer} shows that the optimization recipe and the use of register tokens both matter for the final distilled backbone. Replacing AdamW~\citep{loshchilov2017decoupled} with LARS~\citep{you2017large} improves AP from 54.0 to 54.3, indicating that LARS provides a better optimization regime for matching compact student features to a stronger teacher in this setting. The same table also shows that register tokens~\citep{darcet2023vision} are beneficial: removing them lowers AP to 53.8, while adding one register restores performance to the best result. Increasing the number of registers beyond one does not produce further gains, with both two and four registers reaching 54.2 AP. We therefore adopt LARS together with a single register token as the default configuration, since this combination gives the best accuracy while keeping the student backbone compact.

\subsubsection{Detection Analysis}

\paragraph{Ablation study on patch embedding.}
We evaluate the patch embedding design in a unified ablation that compares the vanilla large-stride embedding with the proposed convolutional stem, and further varies the dilation rate of the final stem convolution to control the receptive field. As shown in Table~\ref{tab:patch_embed}, the convolutional stem slightly increases computation over the vanilla embedding but yields a clear AP gain, especially on small objects. Increasing the dilation rate beyond the default setting degrades performance, particularly on small and medium objects, indicating that a progressively enlarged but still moderate receptive field is more suitable for dense localization.

\begin{table}[!t]
\centering
\caption{\textbf{Effect of patch embedding design and stem receptive field.} \textbf{Vanilla} uses a single $16 \times 16$ stride-16 convolution. \textbf{ConvStem} follows the convolutional-stem design used in the final model and consists of four $3 \times 3$ stride-2 convolutions~\citep{xiao2021early}. Here, $d$ denotes the dilation rate of the final stem convolution, which controls the effective receptive field of the patch embedding stage. All results are reported on ECDet-M.}
\label{tab:patch_embed}

\begin{tabular}{l ccccccc}
\toprule
Patch Embedding & Params (M) & FLOPs (G) & Latency (ms) & AP & AP$_S$ & AP$_M$ & AP$_L$\\
\midrule
Vanilla & 19.0 & 82.5 & 7.87 & 53.5 & 33.7 & 58.2 & 72.8 \\
\rowcolor[gray]{0.95} \textbf{ConvStem ($d{=}1$)} & \textbf{19.2} & \textbf{84.9} & \textbf{7.98} & \textbf{54.3} & \textbf{35.9} & \textbf{59.1} & \textbf{72.7} \\
ConvStem ($d{=}2$) & 19.2 & 84.9 & - & 53.0 & 33.7 & 57.9 & 72.5 \\
ConvStem ($d{=}3$) & 19.2 & 84.9 & - & 53.6 & 33.2 & 58.6 & 72.6 \\
\bottomrule
\end{tabular}
\end{table}

\paragraph{Ablation study on fusion methods.}
Table~\ref{tab:layer_fusion} compares different feature-fusion strategies across the final transformer blocks. Averaging the last two layers offers the best accuracy--efficiency trade-off, while concatenation and the heavier STA fusion used in DEIMv2~\citep{huang2025deimv2} provide no consistent advantage once parameter and FLOP costs are considered.

\begin{table}[!t]
\centering
\caption{\textbf{Effect of layer fusion strategy.} We compare simple mean fusion over different late-layer ranges with concatenation and the STA fusion module from DEIMv2~\citep{huang2025deimv2}. The gray row marks the adopted default setting in ECDet-M.}

\label{tab:layer_fusion}
\small 
\begin{tabular}{l l c c c c}
\toprule
{Fusion Method} & {Fused Layers} & {Params (M)} & {FLOPs (G)} & {Latency (ms)} & {AP} \\
\midrule
Mean & $L_{11}$             & 19.2 & 84.9 & 7.98 & 54.1 \\
\rowcolor[gray]{0.95} \textbf{Mean} & \textbf{$L_{10 \to 11}$}      & \textbf{19.2} & \textbf{84.9} &  \textbf{7.98} & \textbf{54.3}  \\
Mean & $L_{9 \to 11}$       & 19.2 & 84.9 &  7.98 & \textbf{54.3} \\
Mean & $L_{6 \to 11}$  & 19.2 & 84.9 &  8.11 & 54.1 \\
\midrule
Concat & $L_{10 \to 11}$      & 19.5 & 87.0 &  8.05 & 54.2  \\
\midrule
STA~\citep{huang2025deimv2} & $L_{5}$, $L_{7}$, $L_{11}$   & 19.4 & 86.3 & 8.48 &\textbf{54.3} \\
\bottomrule
\end{tabular}
\end{table}

\paragraph{Effect of Task-Specialized Teachers.}
To further assess the impact of task specialization in the teacher, we introduce a strong detection teacher as a complementary baseline. Specifically, we use the Swin-S~\citep{liu2021swin} backbone of Co-DETR~\citep{zong2023detrs}, a state-of-the-art object detector, as the teacher, while keeping the ECDet-S student architecture and overall training budget unchanged. As shown in Table~\ref{tab:teacher}, randomly initialized ECDet-S achieves 46.6 AP, while distillation from the Co-DETR-Swin-S teacher substantially improves the AP to 50.4, demonstrating the effectiveness of a task-specialized detection teacher. Using the DINOv3-based task-specialized ECTeacher-S further raises the AP to 51.7, achieving a 1.0 AP gain over DINOv3-S (50.7 AP). These results demonstrate that task specialization of the teacher can provide substantial gains, while task specialization of the underlying vision model can further enhance the teacher's effectiveness.

\begin{table}[h]
\centering
\caption{\textbf{Comparison of different teacher models.} We additionally evaluate Co-DETR with a Swin-S backbone~\citep{zong2023detrs,liu2021swin} as a strong detection teacher, while keeping the ECDet-S student architecture and overall training budget unchanged.}
\label{tab:teacher}
\begin{tabular}{ccc}
\toprule
Student & Teacher & AP \\
\midrule
\multirow{4}{*}{ECViT-T} & -- & 46.6 \\
& Co-DETR-Swin-S~\cite{} & 50.4 \\
& DINOv3-S & 50.7 \\
& ECTeacher-S & \textbf{51.7} \\
\bottomrule
\end{tabular}
\end{table}

\paragraph{Training cost.}
Table~\ref{tab:gpu_hours} further reports the training cost in GPU hours to provide a more comprehensive view of efficiency beyond the mere number of epochs. Compared to RT-DETRv4-M, ECDet-M achieves a performance gain of \textbf{+0.8 AP} while reducing the total training cost by approximately \textbf{32\%} ($\sim$130 vs. $\sim$190 GPU hours), even when accounting for the additional $\sim$34 GPU hours required for distillation.

\begin{table*}[htbp]
\caption{\textbf{Comparison of training cost on COCO.}
All models are trained on \textbf{4$\times$ NVIDIA RTX 4090 GPUs}.
\textcolor{gray}{Gray results} indicate pre-training on Objects365~\citep{shao2019objects365}
followed by fine-tuning on COCO.}
\label{tab:gpu_hours}
\centering
\begin{tabu}{l c c c c}
\toprule
Method & \#Epochs & Detection & Distillation & Total GPU Hours $\downarrow$ \\
\midrule
\rowfont{\color{gray}}YOLO26-M & 80  & $\sim$24 & -- & $\sim$24 \\
YOLO11-M & 500 & $\sim$133 & -- & $\sim$133 \\
RT-DETRv4-M & 102 & $\sim$190 & -- & $\sim$190 \\
\rowcolor[gray]{0.95}
ECDet-M & 62 & $\sim$96 & $\sim$34 & \textbf{$\sim$130} \\
\bottomrule
\end{tabu}
\end{table*}

\begin{table*}[t]
\begin{minipage}[t]{0.45\linewidth}
\caption{\textbf{Initialization ablation on instance segmentation.}}
\label{tab:insseg_init}
\centering
\begin{tabu}{lcc}
\toprule
Model & Initialization & AP \\
\midrule
ECInsSeg-S & Random & 38.0 \\
ECInsSeg-S & Distilled Backbone & 42.8 \\
ECInsSeg-S & Detection Backbone & \textbf{43.0} \\
\midrule
ECInsSeg-L & Distilled Backbone & 46.4 \\
ECInsSeg-L & Detection Backbone & \textbf{47.1} \\
\bottomrule
\end{tabu}
\end{minipage}
\hfill
\begin{minipage}[t]{0.45\linewidth}
\caption{\textbf{Initialization ablation on human pose estimation.}}
\label{tab:pose_init}
\centering
\begin{tabu}{lcc}
\toprule
Model & Initialization & AP \\
\midrule
ECPose-S & Random & 58.6 \\
ECPose-S & Distilled Backbone & 68.4 \\
ECPose-S & Detection Backbone & \textbf{68.9} \\
\midrule
ECPose-L & Distilled Backbone & 72.5 \\
ECPose-L & Detection Backbone & \textbf{73.5} \\
\bottomrule
\end{tabu}
\end{minipage}
\end{table*}

\paragraph{Initialization Ablation on Downstream Tasks.}

To further investigate the transferability of the proposed backbone, we evaluate different initialization strategies on instance segmentation and human pose estimation. Specifically, we compare random initialization, initialization from the distilled backbone, and initialization from the detection-trained backbone. The results are reported in Table~\ref{tab:insseg_init} and Table~\ref{tab:pose_init}.

For instance segmentation, the distilled backbone provides a substantially stronger initialization than random initialization, improving ECInsSeg-S from 38.0 to 42.8 AP, a gain of \textbf{+4.8 AP}. Initializing with the detection-trained backbone further improves the performance to 43.0 AP. A similar trend is observed with the larger ECInsSeg-L, where the detection-trained backbone improves AP from 46.4 to 47.1 compared with the distilled backbone.

The same trend consistently holds for human pose estimation. As shown in Table~\ref{tab:pose_init}, ECPose-S improves from 58.6 AP with random initialization to 68.4 AP with the distilled backbone, yielding a substantial \textbf{+9.8 AP} improvement. The detection-trained backbone further increases the performance to 68.9 AP. For ECPose-L, the detection-trained backbone also outperforms the distilled backbone, improving AP from 72.5 to 73.5. These results demonstrate that the learned representations provide strong initialization for diverse downstream vision tasks, while the detection-trained backbone offers the most effective initialization among the evaluated strategies.

\section{Conclusion}
We presented \edgecrafter, a unified compact ViT framework for edge dense prediction centered on \detector. The key idea is to adapt a large pretrained ViT into a detection-specialized teacher and distill its representation into compact student backbones designed for edge deployment. Combined with a lightweight convolutional stem and simple multi-scale feature generation, this yields strong accuracy--efficiency trade-offs for object detection under compact parameter and FLOP budgets. The resulting detection-distilled representation also transfers effectively to instance segmentation and human pose estimation through lightweight task-specific prediction modules. Across these tasks, \edgecrafter~achieves strong results on COCO while remaining competitive with, and in several settings outperforming, methods that use additional task-annotation sources such as Objects365 or SAM2 pseudo-masks. Overall, our results indicate that compact ViTs can serve as a practical foundation for edge dense prediction when paired with task-specialized distillation and edge-oriented architectural design.

\section*{Acknowledgments}
We thank the technical staff at Intellindust for their support. We are also grateful to Caizhi Zhu and Xiao Zhou for their help, and to Shihua Huang, Zhe Feng, and Yuting Li for fruitful discussions.

\bibliography{tmlr}
\bibliographystyle{tmlr}

\clearpage
\appendix
\section{Appendix}
\label{sec:appendix}

\subsection{Distillation Pre-training.}
\label{sec:disdetails}
Before downstream task training, we distill the compact ECViT backbones from detection-specialized teachers. Each teacher is constructed by adapting a pretrained DINOv3 model to object detection using the same formulation as ECDet. Specifically, we fine-tune DINOv3-B within the ECDet-X framework to obtain ECTeacher-B, and adapt DINOv3-S with ECDet-L to produce ECTeacher-S. 

Distillation is then performed on a combined dataset comprising ImageNet-1K~\citep{deng2009imagenet} and the COCO training split~\citep{lin2014microsoft}. All student backbones are trained for 50 epochs with LARS~\citep{you2017large}, using a linear warm-up for the first 5 epochs followed by cosine decay to $10^{-3}$ of the peak learning rate. Adopting the scaling rule from~\citet{lightly_train_docs}, the peak learning rate is determined by $\mathrm{base~learning~ rate}\cdot\sqrt{B/1536}$, where $B$ denotes the total batch size. We set the base learnling rate to $4.0$ for ECViT-T/T+ and $9.0$ for ECViT-S/S+, respectively.  We employ a weight decay of $10^{-6}$ and distribute the training across 2$\times$\ NVIDIA RTX 4090 GPUs with a total batch size of $128$. Distillation uses random resized cropping at $224 \times 224$, horizontal flipping, color jitter, grayscale conversion, Gaussian blur, and Mixup augmentation~\citep{zhang2018mixup}. 

\subsection{Implementation Details for ECDet ECPose and ECInsSeg}
\label{sec:appendix_implementationdetails}
\subsubsection{Object Detection}
We provide detailed hyperparameters in Table~\ref{tab:detdetails}. Precisely, for object detection, we train ECDet on the COCO \texttt{train2017} split and evaluate on \texttt{val2017}. All detector variants use an input resolution of $640 \times 640$. We report the standard COCO detection metrics, including AP, AP$_{50}$, AP$_{75}$, AP$_S$, AP$_M$, and AP$_L$. We train the ECDet variants using the AdamW optimizer with a total batch size of 32 on 4$\times$\ NVIDIA RTX 4090 GPUs. The Base LR is fixed at $5 \times 10^{-4}$, while the Backbone LR is specifically scaled down from $2.5 \times 10^{-5}$ to $2.5 \times 10^{-6}$ as model capacity increases to preserve distilled representations. We adopt a progressive training schedule, ranging from 74 epochs (S) to 50 epochs (L/X), all concluding with a 2-epoch fine-tuning stage that disables heavy augmentations. For data augmentation, we employ Mosaic and Mixup with a probability of 0.75 for smaller variants and 1.0 for larger ones, active during the first half of the training duration to facilitate robust feature learning. The weight decay is set to $10^{-4}$, slightly adjusted to $1.25 \times 10^{-4}$ for the L and X variants. 

\paragraph{About latency.} Latency measurement is a non-trivial problem, as it is highly affected by external GPU conditions, particularly thermal behavior and cooling efficiency. Even under identical hardware, variations in GPU temperature can lead to noticeable differences in inference speed. For example, in our X-scale model, a change in GPU idle temperature from 29 °C to 35 °C results in an approximately 0.2 ms difference in measured latency. As a result, latency numbers reported by different users or systems may not be directly comparable. To ensure a fair and controlled evaluation, we measure the latency of all models under the same experimental conditions. Specifically, all experiments are conducted on an NVIDIA T4 GPU with the idle temperature stabilized at 29 °C. We report inference latency with batch size 1 under FP16 precision using TensorRT (v10.6), following the evaluation protocol of D-FINE~\citep{peng2025dfine} and RT-DETR~\citep{lv2024rt}.

\begin{table*}[!t]
\caption{\textbf{Hyperparameters for \detector.}}
\label{tab:detdetails}
\centering
\begin{tabular}{l c c c c}
\toprule
 Setting      & \detector-S & \detector-M & \detector-L & \detector-X \\
\midrule
Optimizer     & AdamW    & AdamW    & AdamW   & AdamW   \\
Backbone LR   & 2.5e-5   & 2.5e-5   & 5e-6    & 2.5e-6  \\
Base LR       & 5e-4     & 5e-4     & 5e-4    & 5e-4    \\
LR decay      & 0.5      & 0.5      & 0.5       & 0.5        \\
Weight Decay  & 1e-4     & 1e-4     & 1.25e-4    & 1.25e-4 \\
\midrule
Epochs (w/ + w/o Aug.)& 72 + 2 & 60 + 2   & 48 + 2    & 48 + 2    \\
Prob$_\text{mosaic}$& 0.75    & 0.75     & 1       & 1        \\
Epochs$_\text{mosaic}$& 36    & 30       & 24      & 24        \\
Prob$_\text{mixup}$   & 0.75  & 0.75     & 1       & 1        \\
Epochs$_\text{mixup}$ & 36    & 30       & 24      & 24        \\
Total Batch Size& 32    & 32       & 32      & 32        \\
\midrule
$\lambda_{\mathrm{cls}}$ & 1       & 1     & 1       & 1        \\
$\lambda_{\ell_1}$       & 5       & 5     & 5       & 5        \\
$\lambda_{\mathrm{giou}}$& 2       & 2     & 2       & 2        \\
$\lambda_{\mathrm{ddf}}$ & 1.5    & 1.5     & 1.5       & 1.5     \\
$\lambda_{\mathrm{fgl}}$ & 0.15    & 0.15     & 0.15       & 0.15        \\

\bottomrule
\end{tabular}
\end{table*}

\subsubsection{Human Pose Estimation}
The training protocol for human pose estimation largely follows the configuration of \detector\ and DETRPose~\citep{janampa2025detrpose}, with several task-specific adjustments summarized in Table~\ref{tab:posedetails}. Specifically, we increase the Total Batch Size to 64. The training duration is extended to 92 epochs for smaller variants (S/M) and 74 epochs for larger variants (L/X), maintaining a 2-epoch final fine-tuning stage without augmentations. For the optimization objective, we employ a combination of classification, keypoint regression, and OKS losses, with the loss weights set to $\lambda_{\mathrm{cls}}=2$, $\lambda_{\mathrm{kpt}}=10$, and $\lambda_{\mathrm{oks}}=4$, respectively.

\begin{table*}[!t]
\caption{\textbf{Hyperparameters for \pose.}}
\label{tab:posedetails}
\centering
\begin{tabular}{l c c c c}
\toprule
 Setting      & \pose-S  & \pose-M  & \pose-L & \pose-X \\
\midrule
Optimizer     & AdamW    & AdamW    & AdamW   & AdamW   \\
Backbone LR   & 2.5e-5   & 2.5e-5   & 2.5e-6    & 2.5e-6  \\
Base LR       & 5e-4     & 5e-4     & 5e-4    & 5e-4    \\
Weight Decay  & 1e-4     & 1e-4     & 1.25e-4    & 1.25e-4 \\
\midrule
Epochs (w/ + w/o Aug.)& 90 + 2 & 90 + 2   & 72 + 2    & 72 + 2    \\
Prob$_\text{mosaic}$  & 0.5  & 0.5     & 0.5     & 0.5      \\
Epochs$_\text{mosaic}$& 45    & 45       & 48  & 48        \\
Prob$_\text{mixup}$   & 0.5  & 0.5     & 0.5       & 0.5       \\
Epochs$_\text{mixup}$ & 45  & 45 & 48  & 48   \\
Prob$_\text{copypaste}$   & 0.5  & 0.5     & 0.5       & 0.5       \\
Epochs$_\text{copypaste}$ & 45  & 45 & 48  & 48   \\
Total Batch Size   & 64    & 64       & 64      & 64        \\
\midrule
$\lambda_{\mathrm{cls}}$ & 2      & 2     & 2       & 2        \\
$\lambda_{\mathrm{kpt}}$& 10      & 10    & 10      & 10        \\
$\lambda_{\mathrm{oks}}$ & 4      & 4     & 4       & 4         \\

\bottomrule
\end{tabular}
\end{table*}
\FloatBarrier
\thispagestyle{fancy}

The data augmentation strategy remains consistent with detection, and we employ a moderated Mosaic and Mixup probability of 0.5.

\subsubsection{Instance Segmentation}
We adopt the same training protocols as \detector, including the optimizer, learning rate schedules, and data augmentations, while keeping the total batch size consistent across 8 GPUs. All \seg~variants use $\lambda_{\mathrm{cls}}=2$, $\lambda_{\ell_1}=\lambda_{\mathrm{giou}}=1$, $\lambda_{\mathrm{ddf}}=1.5$, and $\lambda_{\mathrm{fgl}}=0.15$. Mask prediction is supervised by binary cross-entropy and Dice losses with $\lambda_{\mathrm{mask}}=\lambda_{\mathrm{dice}}=5$.

\end{document}